\documentclass{article}

\usepackage{microtype}
\usepackage{graphicx}
\usepackage{subcaption}
\usepackage{booktabs} 

\usepackage{hyperref}

\usepackage[preprint]{icml2026}

\usepackage{amsmath}
\usepackage{amssymb}
\usepackage{mathtools}
\usepackage{amsthm}
\newcommand{\va}[1]{\mathbf{a}_{#1}}

\usepackage[capitalize,noabbrev]{cleveref}

\theoremstyle{plain}

\theoremstyle{definition}

\theoremstyle{remark}

\begin{document}

\twocolumn[
  \icmltitle{Comparing and Integrating Different Notions of Representational Correspondence in Neural Systems}

  \begin{icmlauthorlist}
  \icmlauthor{Jialin Wu}{cse}
  \icmlauthor{Shreya Saha}{ece}
  \icmlauthor{Yiqing Bo}{cse}
  \icmlauthor{Meenakshi Khosla}{cse,cogs}
  \end{icmlauthorlist}
    
    \icmlaffiliation{cse}{Department of Computer Science and Engineering, UC San Diego, San Diego, CA 92037, USA}
    \icmlaffiliation{ece}{Department of Electrical and Computer Engineering, UC San Diego, San Diego, CA 92037, USA}
    \icmlaffiliation{cogs}{Department of Cognitive Science, UC San Diego, San Diego, CA 92037, USA}
    
    \icmlcorrespondingauthor{Meenakshi Khosla}{mkhosla@ucsd.edu}

  \icmlkeywords{Representation Analysis, Model Typology, Vision Models, Transformer, CNN}

  \vskip 0.3in
]



\printAffiliationsAndNotice{}  

\begin{abstract}
The extent to which different biological and artificial neural systems rely on equivalent internal representations to support similar tasks remains a central question in neuroscience and machine learning. Prior work typically compares systems using a single representational similarity metric, even though different metrics emphasize distinct facets of representational correspondence. Here we evaluate a suite of representational similarity measures by asking how well each metric recovers known structure across two domains: for artificial models, whether procedurally dissimilar models (differing in architecture or training paradigm) are assigned lower similarity than procedurally matched models; and for neural data, whether responses from distinct cortical regions are separated while responses from the same region align across subjects. Across both vision models and neural recordings, metrics that preserve representational geometry or tuning structure more reliably separate this structure than more flexible mappings such as linear predictivity. To integrate these complementary facets, we adapt Similarity Network Fusion, originally developed for multi-omics integration, to combine similarity graphs across metrics. The resulting fused similarity yields sharper separation of procedurally defined model families and, when applied to neural data, recovers a clearer hierarchical organization of the ventral visual stream that aligns more closely with established anatomical and functional hierarchies than single metrics. Overall, this approach reveals which dimensions of representational correspondence recover meaningful structure in models and brains, and how complementary notions of similarity can be integrated.
\vspace{1em}
\end{abstract}

\section{Introduction}
\label{intro}
Understanding when two neural systems, biological or artificial, rely on equivalent internal representations is a central challenge in neuroscience and machine learning. Researchers typically compare representations using a single similarity metric, such as Representational Similarity Analysis (RSA) \citep{rsa}, Centered Kernel Alignment (CKA) \citep{kornblith2019similarity}, or linear predictivity \citep{yamins2014performance}. Crucially, these metrics are not interchangeable. Each metric captures a distinct notion of correspondence, preserving representational geometry, aligning unit-level tuning, or measuring alignment in linearly accessible information, and therefore makes distinct assumptions about which representational transformations should be considered irrelevant. As a result, different metrics can yield qualitatively different conclusions about which systems are similar or distinct\citep{soni2024conclusions, cloos2024differentiable}.

This observation points to a deeper issue: representational similarity is inherently multidimensional. Some aspects of representation may be widely shared across systems, reflecting convergent computational solutions, while others may encode signatures that are specific to particular architectures, training objectives, or biological substrates. As a result, representational comparison remains methodologically fragmented, with different similarity measures often yielding conflicting views of correspondence. This raises a fundamental question: which dimensions of representational correspondence can recover meaningful structure? 

Answering this question requires domain-appropriate criteria. In artificial neural networks, a natural benchmark is \emph{procedural correspondence}: whether a similarity measure distinguishes models trained under different architectures or learning paradigms while grouping procedurally matched models. In neural data, an analogous benchmark is \emph{anatomical–functional organization}: whether responses from distinct cortical regions are separable while responses from the same region align across subjects. Together, these criteria provide principled tests for evaluating which representational similarity measures capture meaningful structure.

Using vision models and visual cortical recordings as a controlled testbed, this work makes three central contributions. First, we provide a systematic evaluation of representational similarity measures capturing distinct dimensions of correspondence (e.g. geometry, unit-level tuning, and linearly accessible information) using principled criteria in both domains. We find that metrics preserving representational geometry or unit-level tuning (e.g., RSA and soft-matching variants \citep{khosla2024soft}) more reliably recover known structure, separating procedurally distinct models and differentiating cortical regions, yet more flexible mappings such as linear predictivity yield weaker discrimination. 

Second, recognizing that these representational dimensions are complementary and that no single metric provides a complete characterization, we introduce an integrative framework based on Similarity Network Fusion (SNF) \citep{wang2014similarity}. By combining similarity graphs derived from multiple metrics, the fused representation sharpens recovery of procedural structure among models and reveals clearer anatomical–functional organization across cortical regions than any individual metric alone.

Finally, we show that integrating complementary similarity dimensions enables the construction of meaningful, data-driven typologies of both artificial models and neural systems. This approach follows empirical traditions in psychology, neuroscience, and genetics~\citep{wang2014similarity, letwin2006combined, echtermeyer2011integrating, mukamel2019perspectives} where researchers identify natural groupings by combining multiple behavioral or molecular indices. For AI models, clustering the SNF-fused similarity matrix reveals groupings that emerge from how systems organize information rather than from prior assumptions about architecture or training tasks. For neural data, the fused similarities expose clear cross-subject structure: responses from the same cortical area form compact clusters that are invariant across subjects, with robust separation between areas. These results demonstrate that integrating across representational dimensions not only improves discriminability, but also supports principled discovery of organization consistent with known anatomical and functional hierarchies.


\begin{figure*}[ht]
\centering
\includegraphics[width=0.8\linewidth]{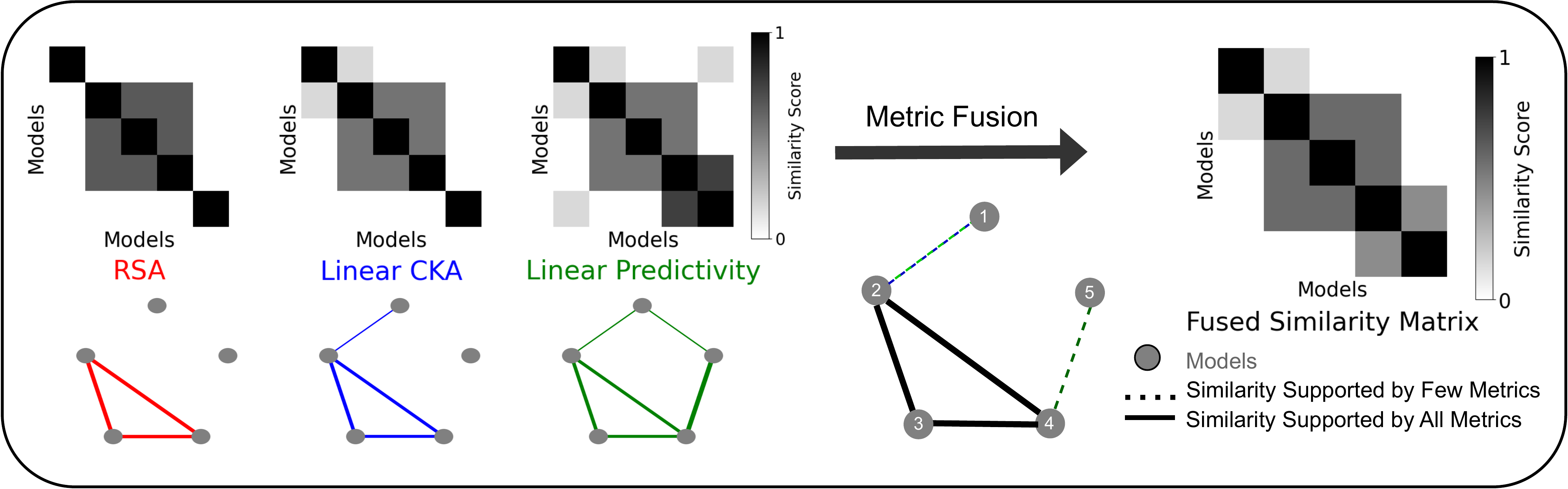}
\caption{
\textbf{Top left:} Each representational metric defines a pairwise similarity matrix over models/brain regions.
\textbf{Bottom left:} Each matrix is visualized as an affinity graph, with nodes representing models/brain regions and edge widths reflecting pairwise similarity strength; weak similarities below a threshold are omitted for clarity.
\textbf{Right:} A consensus matrix obtained via Similarity Network Fusion (SNF) highlights relations consistently supported across metrics while leveraging complementary signals. In the fused graph, solid edges denote agreement across all metrics, dotted edges indicate partial support; strong but uncorroborated edges may persist with reduced weight (e.g., edge 4–5); weak AND metric-specific connections are typically suppressed (e.g., edge 1–5).
}
\label{fig:schematic}
\end{figure*}

\section{Methods}
\label{meth}

\subsection{Model Selection and Dataset}
To evaluate how different representational similarity measures recover known procedural structure, we curated a controlled population of vision models in which architecture and training objective vary systematically while other factors are held fixed. We define a model’s procedure as the combination of its architecture and training paradigm, and treat models sharing the same procedure as belonging to the same model family. We analyze 35 vision models across four primary categories: supervised Convolutional Neural Networks (CNNs), self-supervised CNNs, supervised Transformers, and self-supervised Transformers. We treat ConvNeXt \citep{convnet} and Swin \citep{swin} as distinct families due to their hybrid nature—ConvNeXt incorporates Transformer-inspired design principles within a convolutional architecture, while Swin introduces CNN-like inductive biases into the Transformer framework. We deliberately restricted the model set to encoders pre-trained on ImageNet-1k. This design isolates the effects of architecture and training objective within a shared data and label space so that differences in training data distribution do not impact. For datasets, we use the ImageNet-1k \citep{imagenet_cvpr09} and Ecoset~\citep{ecoset} validation sets and CIFAR10~\citep{CIFAR} and CIFAR100~\citep{CIFAR} test sets. {More} details are provided in Appendix \ref{app:exp} and \ref{app:model}. For brain data, we consider fMRI responses to the 1,000 shared natural images from the Natural Scenes Dataset (NSD)~\citep{nsd}, spanning 10 visual regions across 4 subjects. This dataset enables analogous population-level comparisons in the brain, allowing us to evaluate whether representational similarity measures separate responses from distinct cortical regions while preserving cross-subject alignment within the same region.

\subsection{Representational Metrics}
\label{metr_rep}
We evaluate widely used similarity metrics that differ in the flexibility of the mappings they permit—from permutation-based alignments (soft-matching) to rigid geometric transformations (Procrustes) to looser linear mappings (linear predictivity) as well as non-fitting approaches that compare representational geometry directly (RSA). Consider two representations $\mathbf{X}_i \in \mathbb{R}^{M \times N_i}$ and $\mathbf{X}_j \in \mathbb{R}^{M \times N_j}$ from different models, where $M$ denotes the number of stimuli and $N_i, N_j$ denote the number of units. All representations are mean-centered along the sample dimension. For metrics requiring a fitting procedure (e.g., Soft matching, linear predictivity, Procrustes), similarity values reflect the mean 5-fold cross-validation score.

\textbf{Singular Vector Canonical Correlation Analysis \citep{raghu2017svcca}.} SVCCA first applies singular value decomposition (SVD) to the representation matrices from two models or layers to isolate their most informative directions: $\mathbf{X}_i = \mathbf{U}_i \Sigma_i \mathbf{V}_i^\top$ and $\mathbf{X}_j = \mathbf{U}_j \Sigma_j \mathbf{V}_j^\top$. Retaining the top $N'_i$ and $N'_j$ singular vectors that explain $99\%$ of the variance yields the reduced representations: $\mathbf{X}'_i = \mathbf{U}_i^{(N'_i)\!\top}\mathbf{X}_i$ and $\mathbf{X}'_j = \mathbf{U}_j^{(N'_j)\!\top}\mathbf{X}_j$ whose dominant singular directions capture a disproportionate share of the total information. Canonical correlation analysis (CCA) \citep{hardoon2004canonical} is then applied to these reduced matrices to find linear projections $\mathbf{A}$ and $\mathbf{B}$ that maximize their correlation: $\mathbf{Q} = \max_{\mathbf{A},\,\mathbf{B}} \operatorname{corr}\!\bigl(\mathbf{A}\mathbf{X}'_i,\;\mathbf{B}\mathbf{X}'_j\bigr)$ subject to unit-variance constraints, providing a final similarity score that reflects how closely the subspaces of the two representations align.

\textbf{Projection-Weighted Canonical Correlation Analysis \citep{morcos2018insights}.}
Compared to SVCCA, PWCCA did not apply SVD before CCA and it re-weights the canonical directions according to their contribution to the original representation.
After CCA, we obtain canonical vectors $\mathbf{A}$ and $\mathbf{B}$ and their corresponding correlations $q_i$ for $i=1,2,\ldots,k$, where $k=\min\{N_i,N_j\}$.
Instead of giving each direction equal weight, PWCCA projects the unreduced representation $\mathbf{X}_i$ onto its own canonical vectors to measure how strongly each one reconstructs the data.
The projection weight for the $i^{th}$ canonical direction is $\alpha_i = \frac{\lVert \mathbf{X}_i \va{i} \rVert_{1}}{\sum_{j=1}^{k}\lVert \mathbf{X}_i \va{j} \rVert_{1}},$
where $\va{i}$ is the $i^{th}$ column of $\mathbf{A}$.
The final PWCCA similarity is then the weighted sum of canonical correlations: $\sum_{i=1}^{k} \alpha_i q_i$ which emphasizes directions that explain the greatest fraction of variance in $X_i$ and down-weights noisy, low-variance components, yielding a more faithful measure of representational similarity.

\textbf{Linear Centered Kernel Alignment \citep{kornblith2019similarity}.} Linear CKA provides a scalar measure of how similarly two sets of representations capture the relationships among the same collection of samples. It is is defined as $\frac{\|\mathbf{X}_i^T \mathbf{X}_j\|_F^2}{\|\mathbf{X}_i^T \mathbf{X}_i\|_F \|\mathbf{X}_j^T \mathbf{X}_j\|_F}$, where $\|.\|_F$ is the frobenius norm, and $\mathbf{X}_i$ and $\mathbf{X}_j$ can be assumed to be normalised. Linear CKA is invariant to orthogonal transformations (rotations or reflections), isotropic scaling, and translations of the representations, so it captures only the relational structure shared between the two spaces. 

\textbf{Representational Similarity Analysis ~\citep{rsa}.}
Representational Similarity Analysis (RSA) compares the geometry of representations via their Representational Dissimilarity Matrices (RDMs). For each representation, we compute pairwise dissimilarities between stimuli using $1-\text{Pearson correlation}$, yielding an $M \times M$ RDM that encodes the relational structure. Model similarity is then quantified as the Pearson correlation between their RDMs. RSA is invariant to orthogonal transformations and reflects how models structure their representational spaces.

\textbf{Soft Matching~\citep{khosla2024soft}.}
Soft Matching (SoftMatch) generalizes permutation distance~\citep{ding2021grounding} to representations with different numbers of units by relaxing permutations to ``soft permutations.'' Specifically, consider a non-negative matrix $\mathbf{P} \in \mathbb{R}^{N_i \times N_j}$ whose rows each sum to $1/N_i$ and columns to $1/N_j$. The set of such matrices defines a transportation polytope~\citep{de2013combinatorics}, $\mathcal{T}(N_i, N_j)$. The optimization problem is
\centerline{
$d_T(\mathbf{X}_i, \mathbf{X}_j) = \min_{\mathbf{P} \in \mathcal{T}(N_i,N_j)} \sum_{k,l} \mathbf{P}_{kl}\,\lVert x_i^{(k)} - x_j^{(l)} \rVert^2,$
}
where $x_i^{(k)}$ and $x_j^{(l)}$ are the $k$-th and $l$-th columns (units) of $\mathbf{X}_i$ and $\mathbf{X}_j$. The optimal transport plan $\mathbf{P}^\star$ is found via the network simplex algorithm. When $N_i=N_j$, this reduces to an optimal permutation. The final similarity score is the mean unit-wise correlation between $\mathbf{X}_j$ and $\mathbf{X}_i\mathbf{P}^\star$.

\textbf{Procrustes Alignment~\citep{ding2021grounding}.}
Procrustes analysis finds the orthogonal transformation that best aligns two representations while preserving geometry. For unequal dimensions, the smaller representation is zero-padded. The optimization problem is
$
\min_{\mathbf{R} \in \mathcal{O}(N)} \|\mathbf{X}_j - \mathbf{X}_i\mathbf{R}\|_2^2,
$
where $\mathcal{O}(N) = \{\mathbf{R} \in \mathbb{R}^{N \times N} : \mathbf{R}^\top \mathbf{R} = \mathbf{I}\}$. The optimal transformation $\mathbf{R}^\star$ is obtained via singular value decomposition. The similarity score is the mean unit-wise correlation between $\mathbf{X}_j$ and $\mathbf{X}_i\mathbf{R}^\star$.

\textbf{Linear Predictivity~\citep{yamins2014performance}.}
Linear predictivity seeks an unconstrained linear transformation that best predicts one representation from another: $
\min_{\mathbf{L}} \|\mathbf{X}_j - \mathbf{X}_i\mathbf{L}\|_2^2
$. 
The optimal mapping $\mathbf{L}^\star$ is estimated via ordinary least squares. The final similarity score is the mean unit-wise correlation between $\mathbf{X}_j$ and $\mathbf{X}_i\mathbf{L}^\star$.

\textbf{Average Baseline.} To provide a baseline that naively uses all metrics' information, we symmetrized and min-max rescaled all metrics' matrices and simply averaged them.


\subsection{Separability Metrics}
\label{subsec:sep_metrics}

We next describe the measures used to assess how well representational similarity metrics recover procedural distinctions among models (i.e. how well they separate different model families). 

\textbf{Contrastive Ratio.}
The contrastive ratio quantifies the relative separation between intra-family and inter-family similarities. We consider the similarity values between different models within the same family and take the average of them to obtain $\mu_\text{within}$, and the similarities between models from two model families and obtain the average inter-family $\mu_\text{between}$. The ratio is then defined as $CR = (\mu_\text{within} - \mu_{\text{between}}) / (\mu_{\text{within}} + \mu_{\text{between}})$. A value approaching 1 suggests strong within-family coherence relative to cross-family similarity; a value approaching 0 suggests no difference, and a negative value implies that inter-family similarity exceeds intra-family similarity.

\textbf{D-Prime~\citep{bo}.}
Similarly to contrastive ratio but considering variances, the D-Prime ($d'$) also quantifies the separation between intra-family and inter-family similarities. It is defined as ${(\mu_{\text{within}} - \mu_{\text{between}})}/{\sqrt{0.5(\sigma^2_{\text{within}} + \sigma^2_{\text{between}})}}$,
where $\mu$ and $\sigma^2$ denote the mean and variance of the respective distributions. Higher values indicate tighter clustering within a family and greater spread across families.

\textbf{Silhouette Score~\citep{silhouette}.}
For each model $i$, we compute the average distance $a(i)$ to all other models in the same family and the average distance $b(i)$ to models in the other family. The silhouette value is then $s(i) = ({b(i) - a(i)})/{\max\{a(i), b(i)\}}, s(i) \in [-1, 1]$.
Values near $1$ indicate that the model is well grouped with its own family, values near $0$ suggest boundary placement, and negative values imply greater similarity to another family. The overall silhouette score is obtained by averaging $s(i)$ across models.
\subsection{Similarity Network Fusion}
\label{subsec:snf_desc}

Next, we sought to reconcile the results across the different evaluation metrics. As elaborated in Section~\ref{subsec:sep_metrics}, each metric captures distinct aspects of model representations and varies in its ability to differentiate between model families. To integrate these metrics, we adopt a unified approach inspired by Similarity Network Fusion~\citep{wang2014similarity, snfpy}.
Let $n$ be the number of models,  and $\mathcal{V}$ be the set of the representational metrics. For each metric $v\in\mathcal{V}$, we can get a similarity matrix $\mathbf{S}^v \in \mathbb{R}^{n \times n}$, where each entry $\mathbf{S}^v_{ij}$ measures the similarity between the model $i$'s representation and model $j$'s according to metric $v$, as described in Section~\ref{metr_rep}.  Then, for each metric $v$, we first convert pairwise scores into a dissimilarity matrix~$\mathbf{Q}^v$ as this equation, $Q^v_{ij}=1_\mathrm{i\neq j}\!\left(1-(S^v_{ij}+S^v_{ji})/{2}\right)$. We then build an affinity $\mathbf{W}^v\in\mathbb{R}^{M\times M}$ with a exponential kernel:
\vspace{0.5em}
\newline 
\centerline{$\mathbf{W}^v(i,j) = \frac{1}{\sqrt{2\pi(\sigma^v_{ij})^2}} \exp\left(-\frac{(\mathbf{Q}^v_{ij})^2}{2(\sigma^v_{ij})^2}\right),$}

\vspace{0.5em}
\centerline{$\sigma^v_{ij} = \mu\cdot\frac{\overline{\mathbf{Q}}^v(i,N_i)+\overline{\mathbf{Q}}^v(j,N_j)+\mathbf{Q}^v_{ij}}{3}$.}  

Here, $\overline{\mathbf{Q}}^v(i,N_i)$ denotes the average dissimilarity from $i$ to its $K$ nearest neighbors $N_i$ under metric $v$. We set the hyperparameter $\mu\in(0,1)$ to $0.5$ and $K$ to 5 following the original paper.

We view each $\mathbf{W}^v$ as a weighted graph and aim to fuse them into a single matrix that emphasizes relationships consistently supported across metrics while suppressing spurious ones. Following the implementation, we form a row-normalized full matrix and a KNN-sparse matrix for each metric: 
\vspace{0.25em}\newline
\centerline{
$\mathbf{C}^v_{ii} = \sum_{j} \mathbf{W}^v_{ij}$, 
$\widetilde{\mathbf{W}}^{\,v} = (\mathbf{C}^{v})^{-1}\mathbf{W}^v$}, 

\vspace{0.25em}
\centerline{$\widehat{\mathbf{W}}^{\,v} = \tfrac12(\widetilde{\mathbf{W}}^{\,v} + (\widetilde{\mathbf{W}}^{\,v})^\top)$, 
}

\vspace{0.25em}
\centerline{$\mathbf{S}^{v}_{ij} = \widehat{\mathbf{W}}^{\,v}_{ij} / \sum_{k\in N_i} \widehat{\mathbf{W}}^{\,v}_{ik}$ , if $j \in N_i$; \quad else $0$.}

We then run the SNF message-passing updates with diagonal regularization, which keeps self-affinity dominant while improving numerical stability. Initialize $\mathbf{P}^{(v)}_0=\widehat{\mathbf{W}}^{\,v}$. For $t=0,\dots,T-1$:

\vspace{0.25em}
\centerline{$\mathbf{P}^{(v)}_{t+1} = \mathcal{B}_\alpha\!\left( \mathbf{S}^{(v)} \left(\tfrac{1}{|\mathcal{V}|-1}\sum_{u\neq v}\mathbf{P}^{(u)}_{t}\right) \mathbf{S}^{(v)\top} \right)$} 

\vspace{0.25em}
\centerline{$\mathcal{B}_\alpha(\mathbf{X}) = \tfrac{1}{2}(\mathbf{X}+\mathbf{X}^\top) + \alpha\mathbf{I}$}

\vspace{0.25em}
After $T$ iterations, we average the networks and perform a row normalization and symmetrization: 

\vspace{0.25em}
\centerline{$\mathbf{P} = \frac{1}{|\mathcal{V}|}\sum_{v\in \mathcal{V}}\mathbf{P}^{(v)}_{T}$,  \quad $\mathbf{D}_{ii} = \sum_{j} \mathbf{P}_{ij}$}

\vspace{0.25em}
\centerline{$\widetilde{\mathbf{P}} =  \mathbf{D}^{-1}\mathbf{P}$, \quad
$\widehat{\mathbf{P}} = \tfrac{1}{2}\big(\widetilde{\mathbf{P}}+\widetilde{\mathbf{P}}^\top+\mathbf{I}\big)$.}

Lastly, to form a dendrogram for model typology, we cluster the fused affinity $\widetilde{\mathbf{P}}$ with hierarchical clustering using SciPy’s linkage function~\citep{virtanen2020scipy}.

\section{Results}
\subsection{Which similarity metrics recover procedural differences between models?}
To evaluate which dimensions of representational correspondence recover meaningful structure among models, we systematically compared a suite of representational similarity measures using the procedural criteria introduced above. Our goal is to assess how effectively each metric recovers known procedural structure, distinguishing models that differ in architecture or training paradigm while grouping procedurally matched models.

We quantify recovery of procedural structure using the separability measures described in Methods, which contrast within-family and across-family similarities. Results reported in this section are based on ImageNet representations; qualitatively similar patterns are observed across additional datasets (Appendix~\ref{app:sep_on_other_dataset}). Across metrics, we observe substantial variation in the degree to which procedural structure is recovered (Figures~\ref{fig:ImageNet100_compact_heatmap},~\ref{fig:Ecoset_compact_heatmap},~\ref{fig:CIFAR10_compact_heatmap},~\ref{fig:CIFAR100_compact_heatmap}), indicating that different representational facets carry markedly different information about model families.

Metrics that preserve representational geometry or unit-level tuning properties demonstrate the strongest ability to discriminate between model families. RSA, which captures representational geometry, achieves the highest separability with a $d'$ of 3.95 and silhouette coefficient of 0.56, indicating that geometric organization or relational structure, i.e. how models arrange points in representation
space, constitutes a strong family-specific signature. Linear CKA also shows strong discrimination ($d'$ = 3.91), which aligns with recent theoretical work showing that centered RSA and linear CKA are mathematically equivalent when appropriate centering is applied~\citep{williams2024equivalence}. Similarly, SoftMatch, which preserves individual unit tuning while mapping two representations, shows robust discrimination ($d'$ = 3.59, silhouette = 0.30). Even supervised and self-supervised variants within the same architecture family (particularly CNNs) are reliably separated by these metrics, demonstrating that the training paradigm fundamentally shapes the geometry and tunings of individual neurons, such that metrics diagnostic of these representational facets achieve good separation.  These properties thus constitute the unique representational ``fingerprints'' of model families.

Interestingly, Procrustes alignment, which allows orthogonal transformations, shows intermediate discrimination ($d'$ = 2.96), falling between SoftMatch and Linear Predictivity. This reveals a clear pattern among mapping-based metrics: discriminability decreases monotonically as the transformations become more flexible (SoftMatch $>$ Procrustes $>$ Linear Predictivity). The constraints imposed by less flexible mappings appear to preserve family-specific signatures that are lost by arbitrary linear transformations. 

\begin{figure*}[ht]
\centering
\includegraphics[width=0.9\linewidth]{figures/compact_heatmap_barplot_all_ImageNet100.pdf}
\caption{Model-family separability on ImageNet under $d'$, silhouette score and contrastive ratio. Columns correspond to nine similarity metrics, including two fusion-based methods (SNF, average) and seven commonly used representational metrics {(Distinct aspects of representation emphasized by each metric are shown in the bracketed text).} {The barplot on the right side of each row shows the mean model-family separability. Scores are shown in their native scales and are not directly comparable across measures.}}
\label{fig:ImageNet100_compact_heatmap}
\end{figure*}

\begin{figure*}[ht]
\centering
\includegraphics[width=0.9\linewidth]{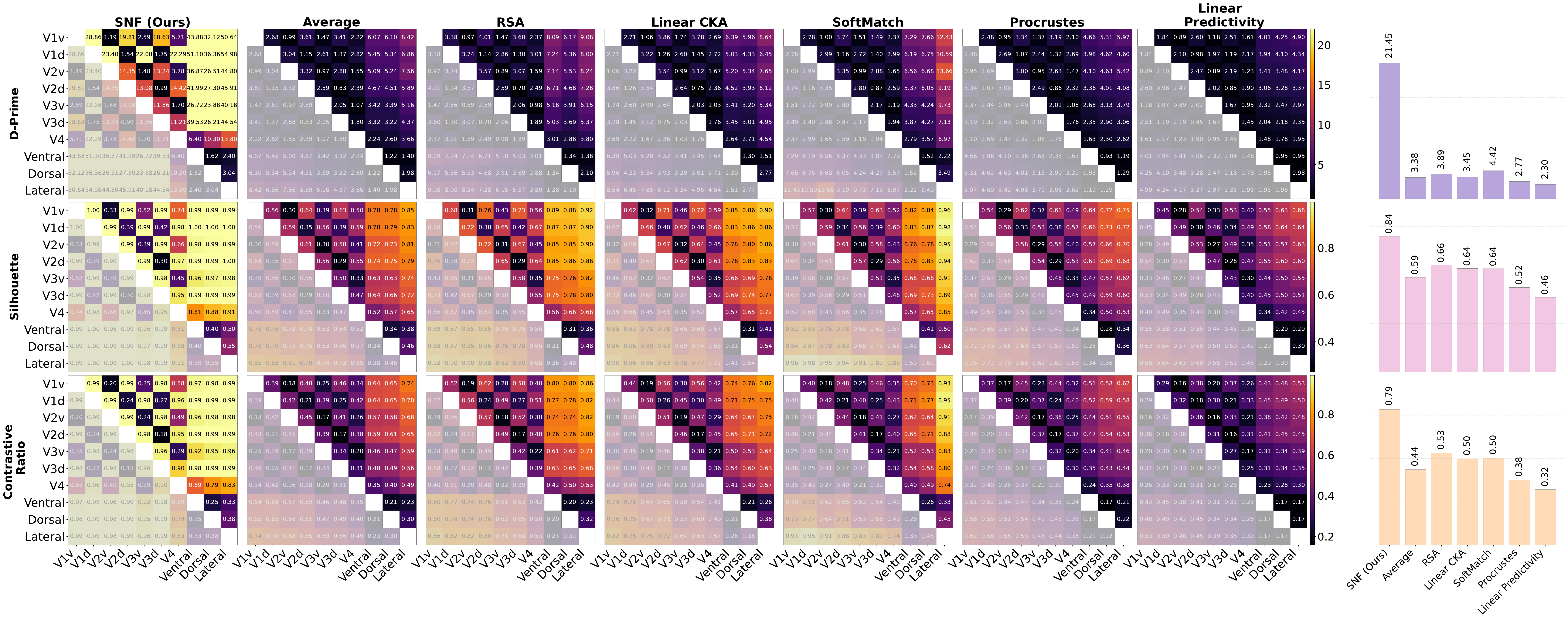}
\caption{Similar to \cref{fig:ImageNet100_compact_heatmap} but on the NSD dataset with human responses and excluding SVCCA and PWCCA.} 
\label{fig:nsd_heatmap_barplot}
\end{figure*}

In contrast, metrics capturing linearly accessible information show substantially weaker discrimination between families. Linear Predictivity demonstrates the lowest separability among direct mapping-based metrics ($d'$ = 2.09, silhouette = 0.14), while CCA-based metrics (PWCCA: $d'$ = 1.55; SVCCA: $d'$ = 1.02) show even weaker family separation. The weak discrimination of CCA-based metrics is particularly revealing. CCA identifies maximally correlated linear projections between representations, finding shared subspaces that are invariant to invertible linear transformations. CCA loads on the linear-accessibility facet: it detects shared linearly decodable subspaces but, unlike RSA/CKA or Procrustes/SoftMatch, it does not constrain or preserve representational geometry or tuning. The invariance of CCA to linear transformations, which makes it powerful for finding shared structure across superficially dissimilar representations, also makes it insensitive to the geometric and topological features that distinguish families. 

The theoretical relationships among these metrics help explain the discrimination hierarchy. RSA and Linear CKA are mathematically equivalent under appropriate centering~\citep{williams2024equivalence} and both preserve the geometric structure of representations—they compare how similarly models organize their representation spaces without fitting any transformation. In contrast, the mapping-based metrics show decreasing discrimination as they allow increasingly flexible transformations: SoftMatch permits only permutations that preserve individual unit correspondences, Procrustes allows orthogonal transformations, while CCA searches for optimal linear projections that maximize correlation. Linear Predictivity provides the most flexibility, allowing any linear transformation that minimizes prediction error.  

The weak discrimination of metrics that assess linearly accessible information might suggest that this aspect of representation is more consistent across model families than geometric organization. Whether this reflects convergent computational strategies, methodological limitations of these metrics, or task-imposed constraints remains open.

\subsection{Integration achieves superior discrimination}\label{sec:snf}

Critically, our SNF approach, which integrates information across all representational dimensions, achieves dramatically superior family separation compared to any single metric. SNF attains a $d'$ of 12.42—nearly three times higher than the best-performing single measure—and consistently outperforms all baselines across separation criteria. Importantly, as shown in Figure~\ref{fig:ImageNet100_compact_heatmap}, SNF maintains high and balanced discrimination across nearly all family pairs. By contrast, individual metrics often exhibit uneven performance, separating some families while failing or for others.

Averaging similarities across metrics does not resolve this limitation: simple means dilute complementary signals and retain conflicting noise. In contrast, SNF’s diffusion-based fusion reinforces consistent neighborhood structure across metrics while attenuating discordant components, yielding both stronger global separation and greater local stability. 

This superior performance demonstrates that different representational dimensions provide complementary information about model families. While geometry and tuning capture family-specific computational strategies, and linearly accessible features potentially reflect more universal solutions, the integration of these diverse facets yields comprehensive signatures that most reliably distinguish model families. 

To test whether SNF recovers the shared structure across metrics rather than replicating any single one, we try to quantify the inter-metric agreement. Specifically, to ensure comparability across metrics, we symmetrize every similarity matrix by averaging it with its transpose, remove the diagonal (self-similarity), vectorize the remaining upper-triangle entries, and then compute correlation between vectors. As shown in Figure~\ref{fig:cross_metrics}, geometry-preserving metrics (RSA, SoftMatch, CKA) show strong mutual agreement, mapping-based metrics (Procrustes, Linear Predictivity) are moderately aligned, and CCA-variants highly correlate with each other but less agree with other metrics. SNF aligns only moderately with any single metric and is clearly distinct from simple averaging, indicating it fuses complementary facets instead of collapsing to one metric. 

\subsection{Which similarity metrics recover known anatomical–functional structure in visual cortex?}
We next evaluate representational similarity metrics using neural data from the Natural Scenes Dataset (NSD), asking whether the patterns observed in artificial models generalize to biological visual systems. We compare representational correspondences across cortical regions using responses to a shared set of natural images, focusing on cross-subject comparisons. Comparisons across different regions within the same subject are excluded due to the spatial blurring inherent to fMRI data, which can artificially inflate within-subject similarity across regions. We exclude SVCCA and PWCCA from the brain analyses, as their alignment values were systematically inflated by the large voxel dimensionality relative to the number of stimuli. To assess recovery of anatomical–functional structure, we quantify region discriminability using the separability measure $d'$, contrasting within-region similarity across subjects with across-region similarity. Across single metrics, we observe patterns consistent with the model results: metrics that preserve representational geometry or unit-level tuning better differentiate cortical regions, whereas metrics with more permissive mappings show weaker discrimination. Critically, integrating information across representational dimensions using Similarity Network Fusion (SNF) yields substantially stronger recovery of anatomical–functional structure than any individual metric. As shown in \cref{fig:nsd_heatmap_barplot}, SNF achieves a mean $d'$ of 21.45—nearly five times higher than the best-performing single measure—and consistently outperforms all baselines across evaluation criteria.

\subsection{Data-driven typologies in models and brains}
Having shown that integrating complementary representational similarity dimensions yields the most informative correspondence structure, we next examine the population-level organization revealed by the SNF-fused similarity matrices. Rather than evaluating individual pairwise comparisons, this analysis asks whether integrated similarity supports the discovery of coherent structure across populations of systems, artificial models and biological brain regions alike, without relying on prior assumptions about architecture, training paradigm, or anatomy.

\subsubsection{Population structure in vision models}
As a baseline, clustering based on individual similarity metrics produces fragmented and inconsistent organization (Figure \ref{fig:metrics_clustering_imagenet}, \ref{fig:metrics_clustering_imagenet_2},
\ref{fig:metrics_clustering_Ecoset_1}, \ref{fig:metrics_clustering_Ecoset_2}, \ref{fig:metrics_clustering_cifar10_1}, 
\ref{fig:metrics_clustering_cifar10_2}
\ref{fig:metrics_clustering_cifar100_1}
\ref{fig:metrics_clustering_cifar100_2}). Metrics such as Linear Predictivity and Procrustes yield relatively uniform similarity values across models, resulting in diffuse clusters with weak separation. SoftMatch, which emphasizes unit-level geometric alignment, struggles to clearly organize models that fall outside standard CNN or supervised regimes. CCA-based metrics (PWCCA, SVCCA) produce noisy similarity matrices in which no stable clustering structure is apparent. RSA reveals partial organization (e.g. a coarse separation between CNNs and Transformers) but the resulting clusters remain diffuse. Together, these results underscore the limitations of single similarity metrics: each emphasizes a particular representational facet, leading to population structures that are incomplete or unstable.

\begin{figure}[bthp]
    \centering
    \includegraphics[width=\linewidth]{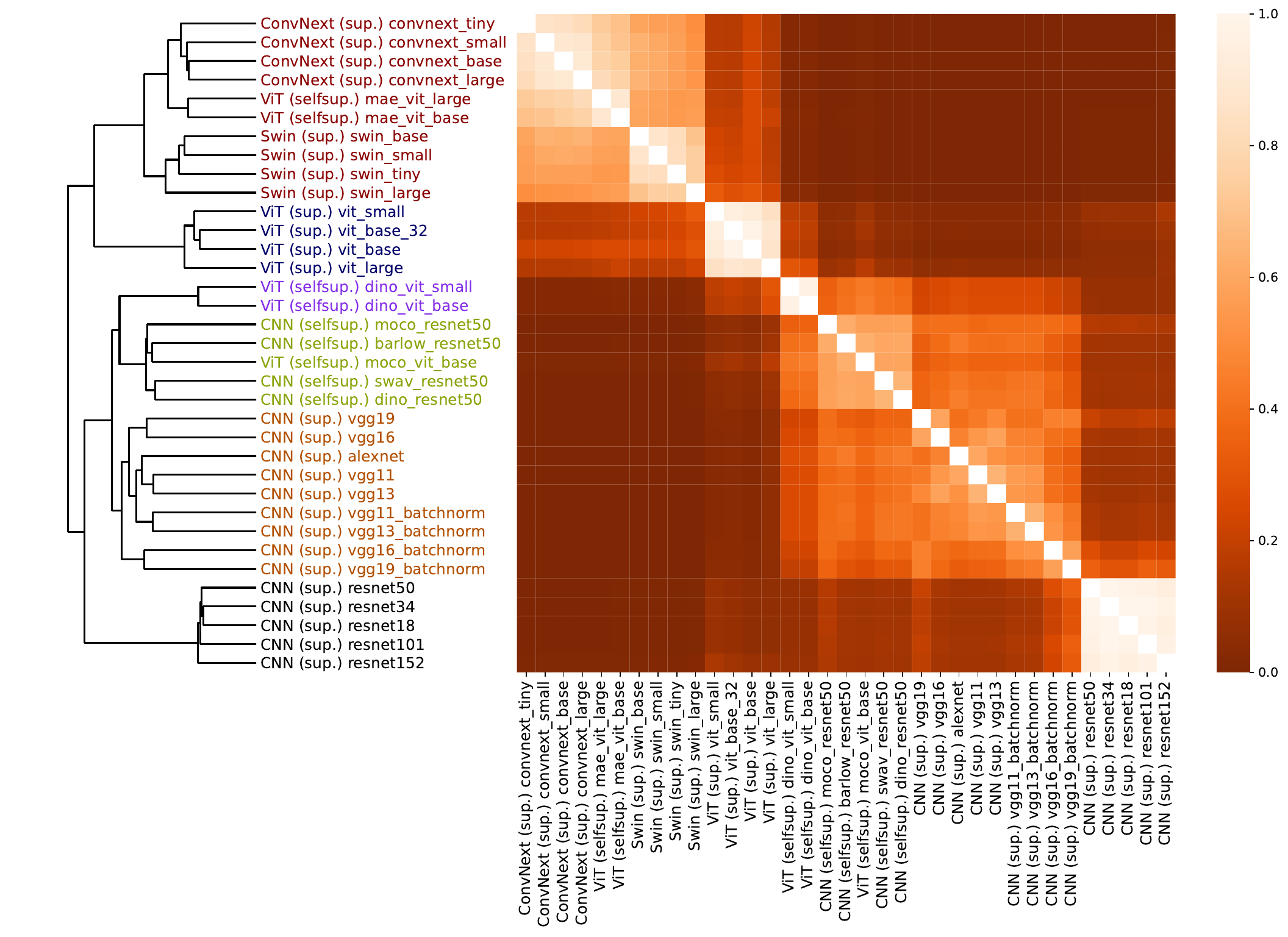}
     \caption{SNF-based clustering reveals that models naturally group by architecture and supervision regime. The heatmap shows the SNF-fused similarity matrix reordered by leaf ordering. Leaf labels are colored by {the cluster (formed by SNF) they belong to.}}
     \label{fig:snf_clustering_ImageNet100}
\end{figure}

In contrast, hierarchical clustering of the fused similarity matrix reveals well-defined groupings that both recover expected relationships and reveal additional organizational structure (Figure \ref{fig:snf_clustering_ImageNet100}, \ref{fig:snf_clustering_Ecoset}, \ref{fig:snf_clustering_cifar10}, \ref{fig:snf_clustering_cifar100})). The fidelity of these clusterings to the underlying similarity structure is quantified using the cophenetic correlation coefficient (Figure~\ref{fig:cophenetic}), which shows that the SNF-based organization more faithfully reflects pairwise similarities than individual metrics.

Examining these clusters reveals a nuanced interplay between architecture and training paradigm. While some groupings align with architectural expectations, such as supervised CNNs and supervised ViTs forming distinct clusters (Figure \ref{fig:snf_clustering_ImageNet100}), the most striking result is that training paradigm can override architectural boundaries. All self-supervised models, regardless of architecture, form a unified cluster that transcends the CNN–Transformer divide. Self-supervised ResNets group more closely with self-supervised ViTs than with their supervised architectural counterparts, suggesting that self-supervised objectives may induce shared computational strategies that dominate over architectural differences. Perhaps most surprisingly, hybrid architectures (ConvNeXt and Swin) cluster with MAE models despite their different design philosophies. ConvNeXts modernize CNNs with Transformer-inspired components, Swins introduce CNN-like biases into Transformers, and MAE employs masked reconstruction - yet they all converge on similar representational structures. This convergence suggests that architectural modernization and masked reconstruction, though distinct in implementation, arrive at related computational solutions.

The typology reveals the ``species'' of vision models, defined not just by their surface characteristics but by how they fundamentally process and organize information.

\subsubsection{Population structure in visual cortex}

\begin{figure*}[htbp!]
  \centering
  \includegraphics[width=0.685\linewidth]{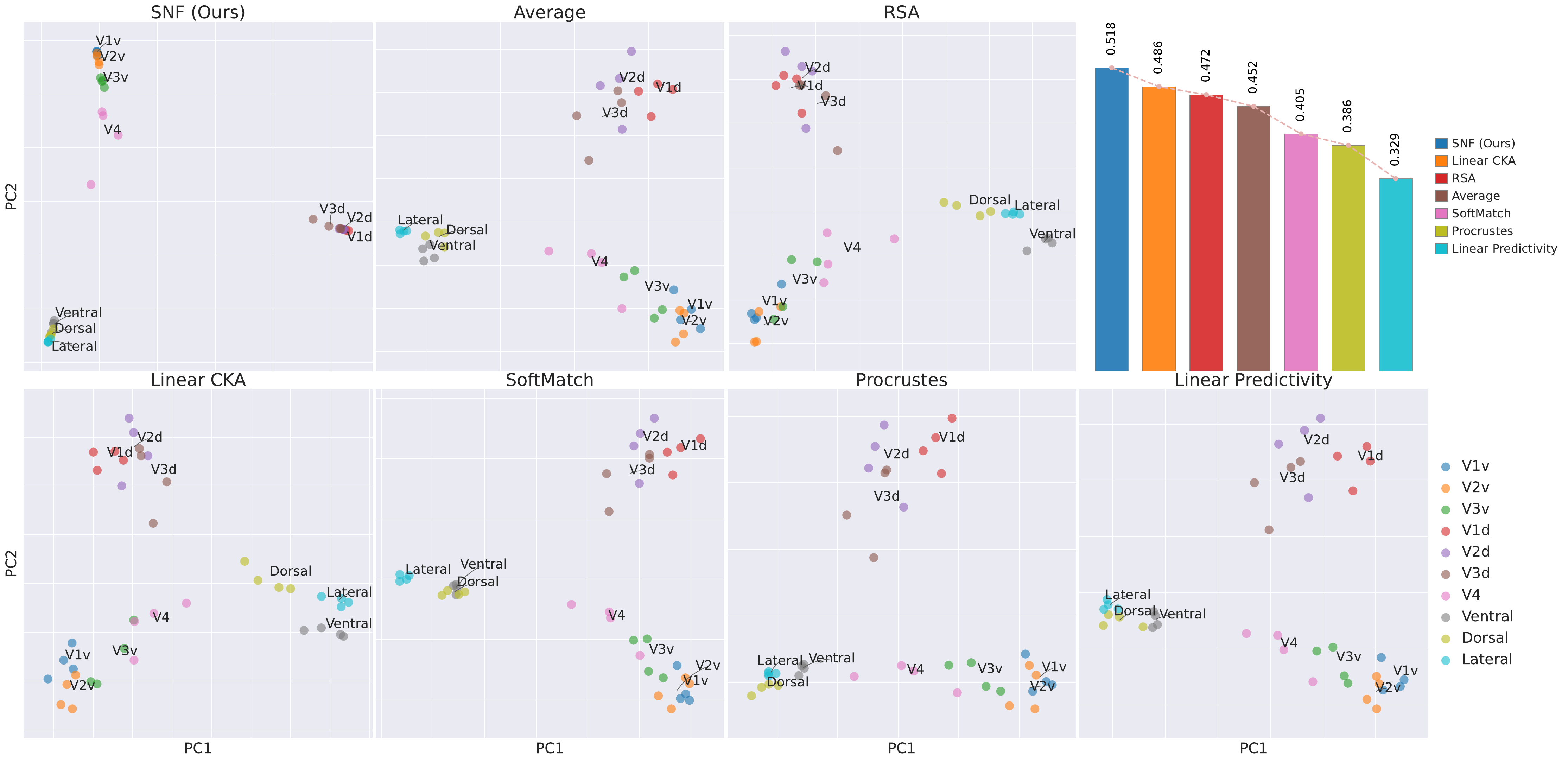}
  \caption{PCA visualization and hierarchial score comparison of cross-brain-region representational relationships in the ventral visual pathway of the NSD Dataset. Each panel corresponds to one metric, points represent brain region instances, and text labels indicate centroid positions. The top-right bar chart displays the raw hierarchical scores for each metric, with colors corresponding to metrics.}
  \label{fig:nsd_hierarchy}
\end{figure*}

We observe a closely parallel pattern in neural data. Applying the same SNF-based clustering approach to representational similarities across visual cortical regions reveals coherent, cross-subject organization consistent with known anatomical and functional structure. Visualization using PCA shows that while individual metrics recover broad trends, the SNF-fused similarity matrix produces the clearest and most interpretable layout, with early visual areas (V1–V4) forming a smooth progression and V4 occupying a transitional position between early and higher-level visual regions (Figure~\ref{fig:nsd_hierarchy}).

To quantify how well a representational similarity measure preserves the expected ventral-stream hierarchy, following \citet{thobani2025modelbraincomparisonusinginteranimal}, we compute a \emph{hierarchy alignment score}. This score measures the rank correlation (Kendall's $\tau$) between representational similarity and approximated hierarchical distance between cortical regions. Each region $r$ is assigned a discrete hierarchical level $L(r) \in \{1,\dots,5\}$, as
$
L(\text{V1v}) = L(\text{V1d})= 1,
L(\text{V2v}) = L(\text{V2d}) = 2,
L(\text{V3v}) = L(\text{V3d}) = 3,
L(\text{V4}) = 4,
L(\text{ventral\_visual}) = 5 
$. Given a brain-region-by-brain-region similarity matrix $S \in \mathbb{R}^{R \times R}$ and the corresponding region labels
$\{r_1,\dots,r_R\}$, we first construct a hierarchy-distance matrix,
$
H_{ij} = \bigl|L(r_i) - L(r_j)\bigr|, \quad 1 \le i,j \le R.
$
We then extract all off-diagonal entries of $H$ and $S$,
$
\mathbf{h} = \{ H_{ij} : i \ne j \}, \quad
\mathbf{s} = \{ S_{ij} : i \ne j \},
$
and define $\mathbf{y} = -\mathbf{s}$ so that smaller hierarchy distance corresponds to larger similarity. The hierarchy score is the Kendall's $\tau$-b rank correlation between $\mathbf{h}$ and $\mathbf{y}$: $\tau_{\text{hier}} = \tau_{\text{Kendall}}\!\bigl(\mathbf{h}, \mathbf{y}\bigr)$.
The results further support the superior performance of SNF in revealing known biological correspondence structure (\cref{fig:nsd_hierarchy}). The ability of individual similarity metrics to recover this hierarchical alignment mirrors the trends observed in the separability analyses: metrics that preserve representational geometry (e.g., RSA, linear CKA) achieve the strongest alignment with the ventral-stream hierarchy, while mapping-based metrics exhibit progressively weaker alignment as the allowable transformations become more flexible, with SoftMatch (which imposes the strictest correspondence) yielding the highest hierarchy alignment among mapping-based approaches.

\section{Discussion}
\label{sec:dis}
This work reframes representational comparison as a question of \emph{what kinds of structure different similarity metrics are capable of recovering}. Rather than treating representational similarity as a single quantity, we systematically evaluated multiple correspondence dimensions (e.g. geometric organization, unit-level tuning, and linearly accessible information) and assessed how well each recovers known structure in two domains: procedural differences among artificial vision models and anatomical–functional organization in human visual cortex. Across both domains, we find a consistent ordering of representational similarity measures. Metrics that preserve representational geometry or unit-level tuning (e.g., RSA, linear CKA, SoftMatch) most reliably recover known structure, distinguishing procedurally distinct model families and differentiating cortical regions across subjects. In contrast, more flexible mappings that emphasize linearly accessible information (e.g., Linear Predictivity and CCA-based methods) show substantially weaker discrimination. These results suggest that while linearly decodable signals may reflect more broadly shared constraints, finer-grained geometric and tuning structure carries stronger signatures of procedural and anatomical organization.

Another contribution of this work is to show that these representational dimensions are complementary rather than redundant. No single metric provides a complete characterization of representational correspondence. To address this, we adapt Similarity Network Fusion (SNF), originally developed for multi-omics integration, to combine similarity graphs derived from multiple metrics. The fused similarity consistently outperforms individual measures in recovering known structure in both models and brains. We note that SNF provides a stable global summary when different metrics capture complementary aspects of representation, but does not replace individual similarity measures. While the fused similarity is well-suited for identifying broad organizational structure, such as model families or cortical hierarchies, individual metrics are necessary to interpret which representational properties drive the observed similarity. More broadly, our results suggest that representational similarity measures should be interpreted in terms of the structure they are capable of recovering, rather than interchangeable proxies for representational correspondence.

\newpage

\section*{Impact Statement}


This paper presents work whose goal is to advance the field of Machine Learning. There are many potential societal consequences of our work, none of which we feel must be specifically highlighted here.


\bibliography{icml2026_paper}
\bibliographystyle{icml2026}

\newpage
\appendix
\onecolumn
\counterwithin{figure}{section}
\counterwithin{table}{section}
\renewcommand{\thefigure}{\thesection.\arabic{figure}}
\renewcommand{\thetable}{\thesection.\arabic{table}}

\section{Experiment Settings}
\label{app:exp}

\textbf{Datasets.}
\label{datasets}
The chosen datasets are balanced across different classes as shown in Tabel~\ref{tab:dataset}.

\begin{table}[htbp]
  \caption{Per-class and total sample counts for standard evaluation splits.}
  \label{tab:dataset}
  \begin{center}
    \begin{small}
      \begin{sc}
        \begin{tabular}{lccc}
          \toprule
          Dataset & Number of Classes & Samples / Class & Total Samples \\
          \midrule
          ImageNet-1k (valid) & 1{,}000 & 50     & 50{,}000 \\
          Ecoset (valid)      & 565     & 50     & 28{,}250 \\
          CIFAR-10 (test)     & 10      & 1{,}000 & 10{,}000 \\
          CIFAR-100 (test)    & 100     & 100    & 10{,}000 \\
          \bottomrule
        \end{tabular}
      \end{sc}
    \end{small}
  \end{center}
  \vskip -0.1in
\end{table}

\textbf{Models.}
\label{weights}
All models are trained on the ImageNet-1K training set. We obtain pretrained weights from torchvision~\citep{torchvision2016}, Torch Hub~\citep{pytorch}, timm~\citep{rw2019timm}, or the official repositories. Unless noted otherwise, we extract activations from each model’s penultimate layer. For CNNs, which commonly include global average pooling, we use that pooled feature. For ViT-style models, we average non-CLS token embeddings to form the final representation for consistency across architectures.

\section{Model Family and Architecture Choices}
\label{app:model}
We evaluate multiple architectures within each family to capture variation in depth, width, parameter amounts and design choices.

\textbf{Convolutional Neural Network (supervised; CNN (sup.)).}
Bottom-up hierarchies with convolutions and pooling that impose strong local inductive biases. We include AlexNet~\citep{alexnet}, VGG-11/13/16/19 (with/without batch normalization)~\citep{vgg}, and ResNet-18/34/50/101/152~\citep{resnet}.

\textbf{Transformer (supervised; Trans (sup.)).}
Vision Transformers partition images into fixed-size patches and use multi-head self-attention for global interactions~\citep{vit}. We include ViT-S/16, ViT-B/16, ViT-L/16, and ViT-B/32.

\textbf{ConvNeXt~\citep{convnet}.}
A convolutional family inspired by Transformer design (e.g., large-kernel depthwise convolutions, patchified stems, inverted bottlenecks). We use ConvNeXt-Tiny/Small/Base/Large.

\textbf{Swin Transformer~\citep{swin}.}
A hierarchical Transformer with shifted window attention for efficient locality while retaining global context. We use Swin-Tiny/Small/Base/Large.

\textbf{Convolutional Neural Network (self-supervised; CNN ({selfsup.})).}
Methods trained without labels using CNN backbones (ResNet-50). We include MoCo~\citep{moco}, DINO~\citep{dino}, SwAV~\citep{swav}, and Barlow Twins~\citep{barlow}, spanning contrastive and non-contrastive paradigms (momentum contrast, self-distillation, online clustering, and redundancy reduction).

\textbf{Transformer (self-supervised; Trans ({selfsup.})).}
Label-free training with Transformer backbones. We include DINO-ViT-Small/16 and DINO-ViT-Base/16 ~\citep{dino}, MoCo-ViT-Base/16~\citep{moco}, and MAE-ViT-Base/16 and MAE-ViT-Large/16~\citep{mae}.

\newpage
\section{Separation Performance on Other Datasets}
\label{app:sep_on_other_dataset}
\begin{figure}[htbp!]
\centering
\includegraphics[width=\linewidth]{figures/compact_heatmap_barplot_all_Ecoset.pdf}
\caption{Same as \cref{fig:ImageNet100_compact_heatmap}, but using Ecoset instead of ImageNet.}
\label{fig:Ecoset_compact_heatmap}
\end{figure}

\begin{figure}[htbp!]
\centering
\includegraphics[width=\linewidth]{figures/compact_heatmap_barplot_all_CIFAR10_big_size.pdf}
\caption{Same as \cref{fig:ImageNet100_compact_heatmap}, but using CIFAR10 instead of ImageNet.}
\label{fig:CIFAR10_compact_heatmap}
\end{figure}

\begin{figure}[htbp!]
\centering
\includegraphics[width=\linewidth]{figures/compact_heatmap_barplot_all_CIFAR100_big_size.pdf}
\caption{Same as \cref{fig:ImageNet100_compact_heatmap}, but using CIFAR100 instead of ImageNet.}
\label{fig:CIFAR100_compact_heatmap}
\end{figure}

\newpage
\section{Metrics' Similarity Scores Consistency}

\begin{figure}[htbp!]
\centering
\includegraphics[width=\linewidth]{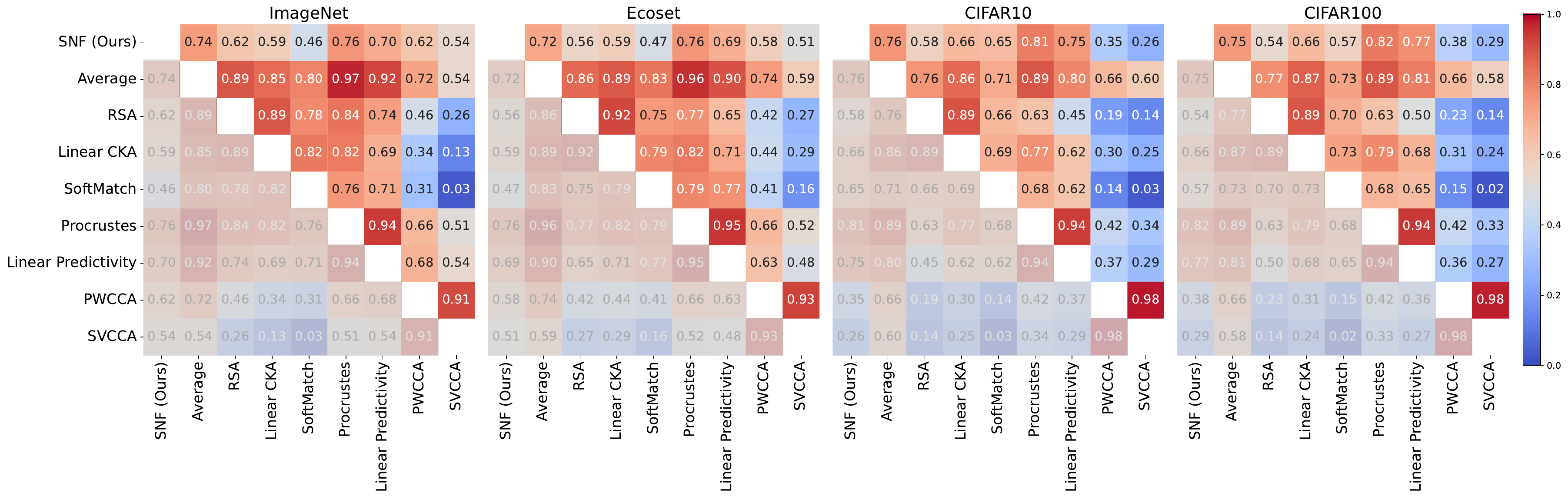}
\caption{Each subplot (ImageNet, Ecoset, CIFAR10, CIFAR100) shows pairwise Pearson correlations between the vectorized upper-triangle entries of the symmetrized model–model similarity matrices produced by nine metrics. Higher values indicate that two metrics have more similar relational geometry among models.}
\label{fig:cross_metrics}
\end{figure}

\newpage
\section{{More Clustering Performance}}

\begin{figure}[ht]
\centering
\includegraphics[width=0.8\linewidth]{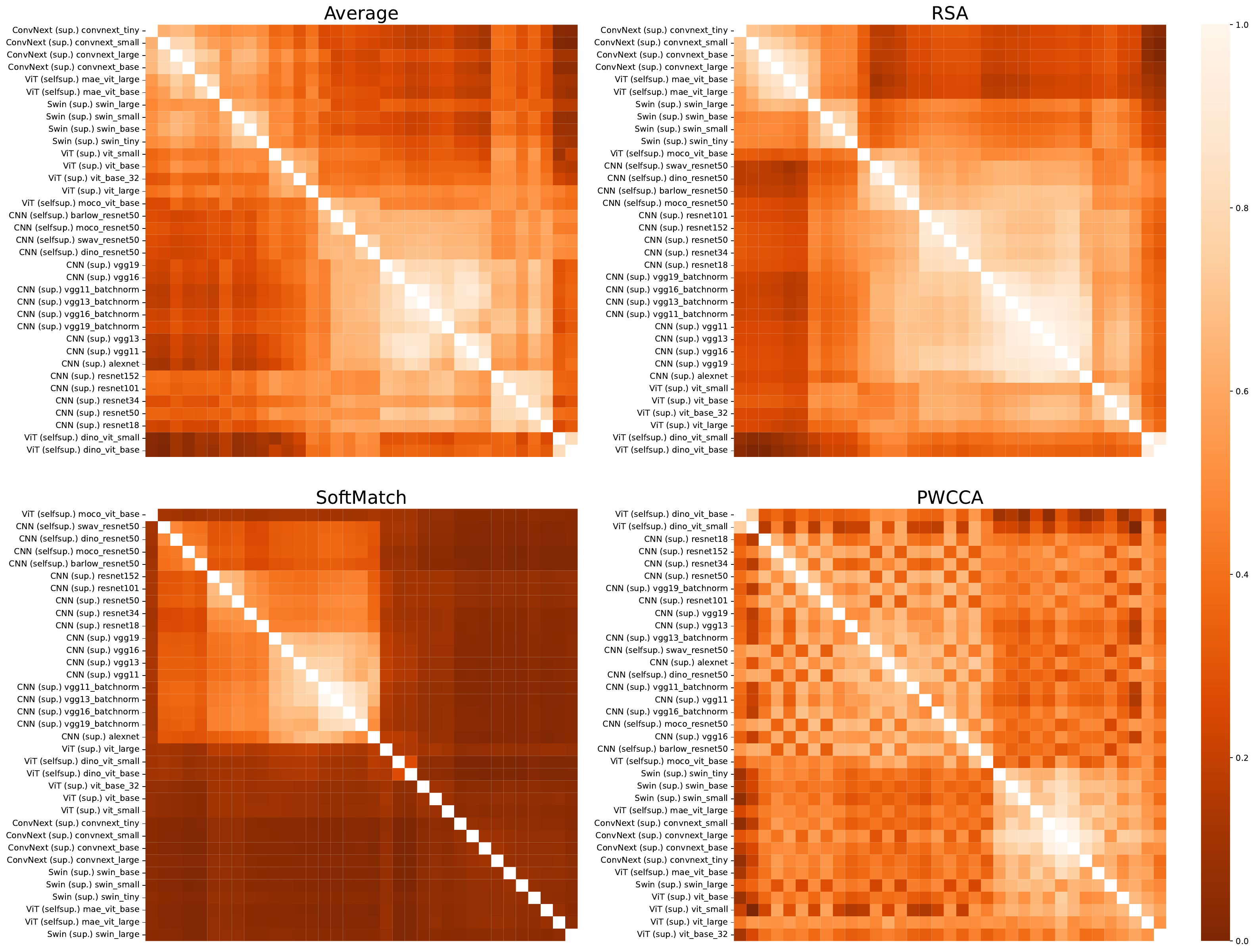}
\caption{\textcolor{black}{Hierarchical clustering of models using {three functionally distinct representational similarity metrics and a similarity-metric–averaging baseline (see Fig.~\ref{fig:metrics_clustering_imagenet_2} for additional metrics). Clustering is performed with average linkage and optimal leaf ordering, based on induced distances ($1 -$ similarity score). Rows and columns are deliberately re-ordered to match the leaf ordering produced by the clustering algorithm.}  Lighter colors indicate higher similarity; diagonal entries (self-comparisons) are omitted.}}
\label{fig:metrics_clustering_imagenet}
\end{figure}

\begin{figure}[htbp!]
\centering
\includegraphics[width=\linewidth]{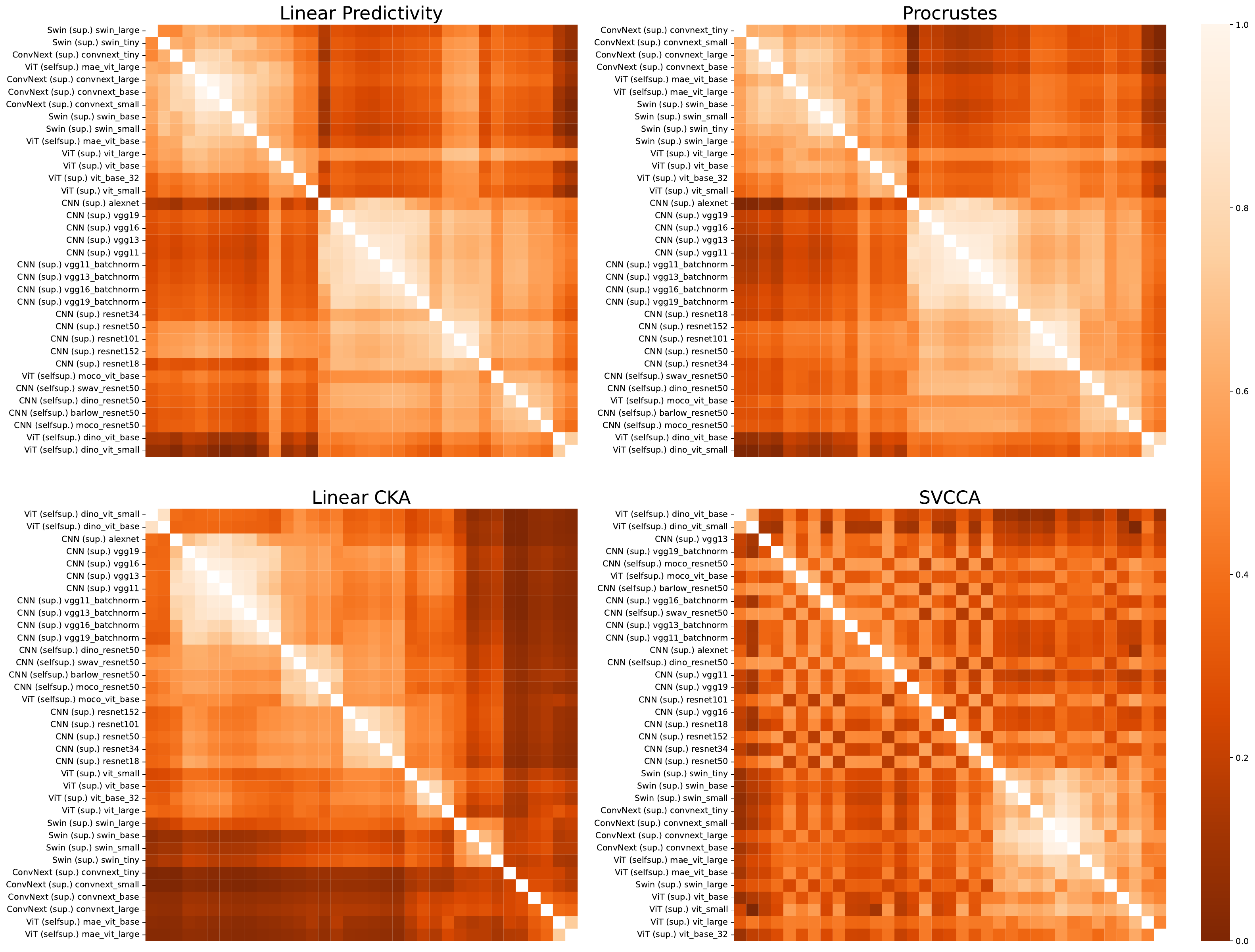}
\caption{Same as \cref{fig:metrics_clustering_imagenet}, but for the other 4 metrics.}
\label{fig:metrics_clustering_imagenet_2}
\end{figure}

\begin{figure}[htbp!]
\centering
\includegraphics[width=0.95\columnwidth]{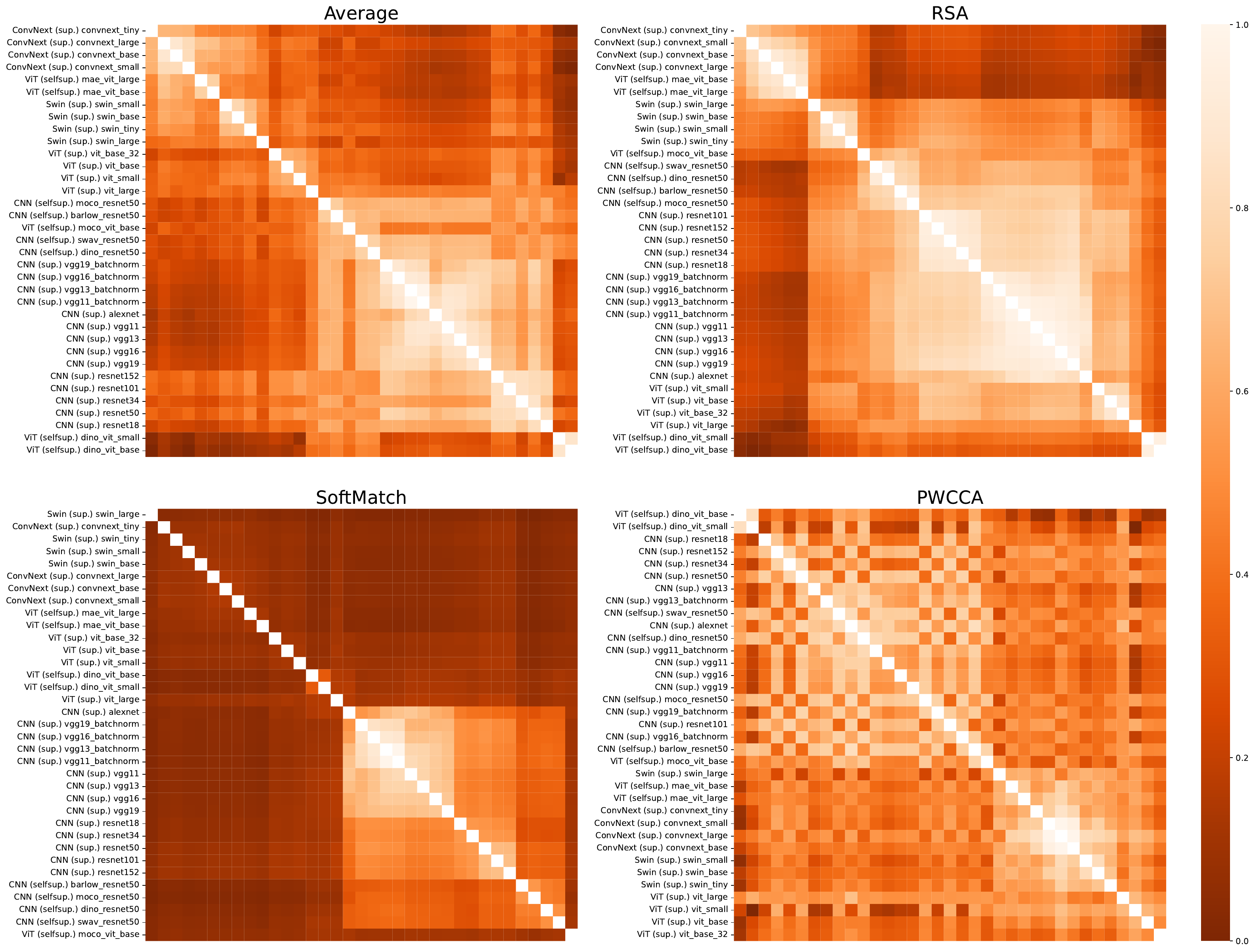}
\caption{Same as \cref{fig:metrics_clustering_imagenet}, but using Ecoset instead of ImageNet.}
\label{fig:metrics_clustering_Ecoset_1}
\end{figure}

\begin{figure}[htbp!]
\centering
\includegraphics[width=0.95\linewidth]{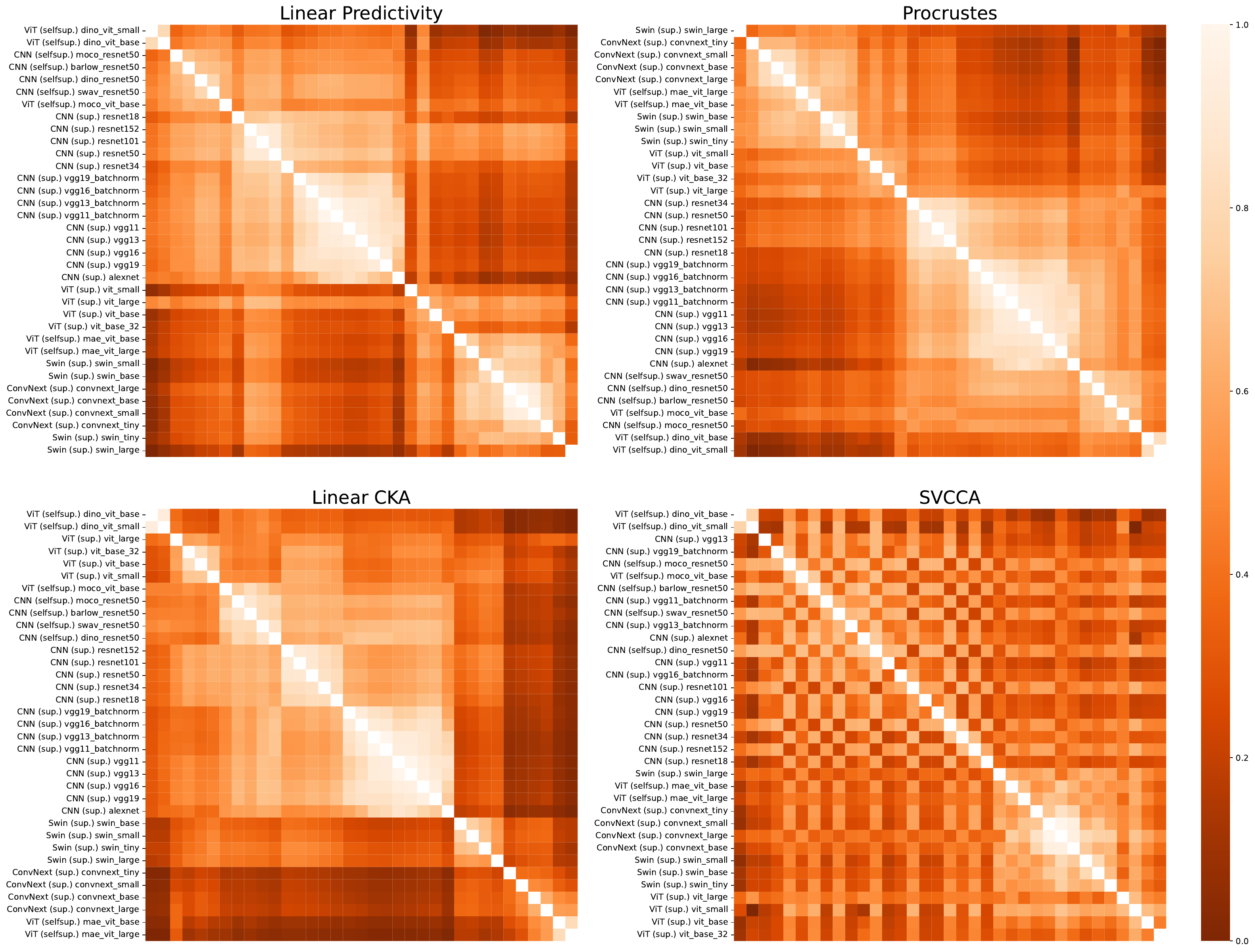}
\caption{Same as \cref{fig:metrics_clustering_imagenet_2}, but using Ecoset instead of ImageNet.}
\label{fig:metrics_clustering_Ecoset_2}
\end{figure}

\begin{figure}[htbp!]
\centering
\includegraphics[width=\linewidth]{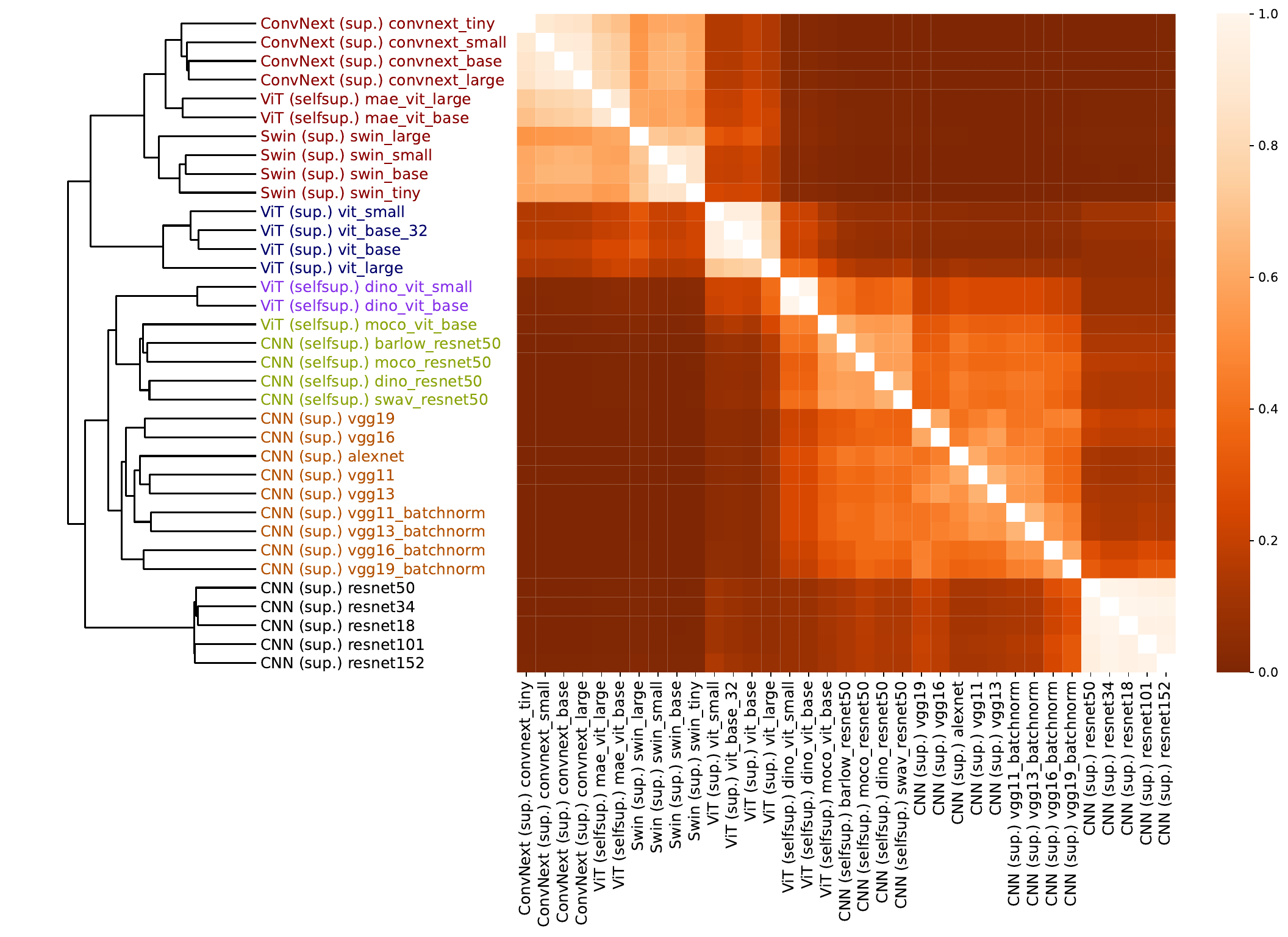}
\caption{Same as \cref{fig:snf_clustering_ImageNet100}, but using Ecoset instead of ImageNet.}
\label{fig:snf_clustering_Ecoset}
\end{figure}

\begin{figure}[htbp!]
\centering
\includegraphics[width=0.95\linewidth]{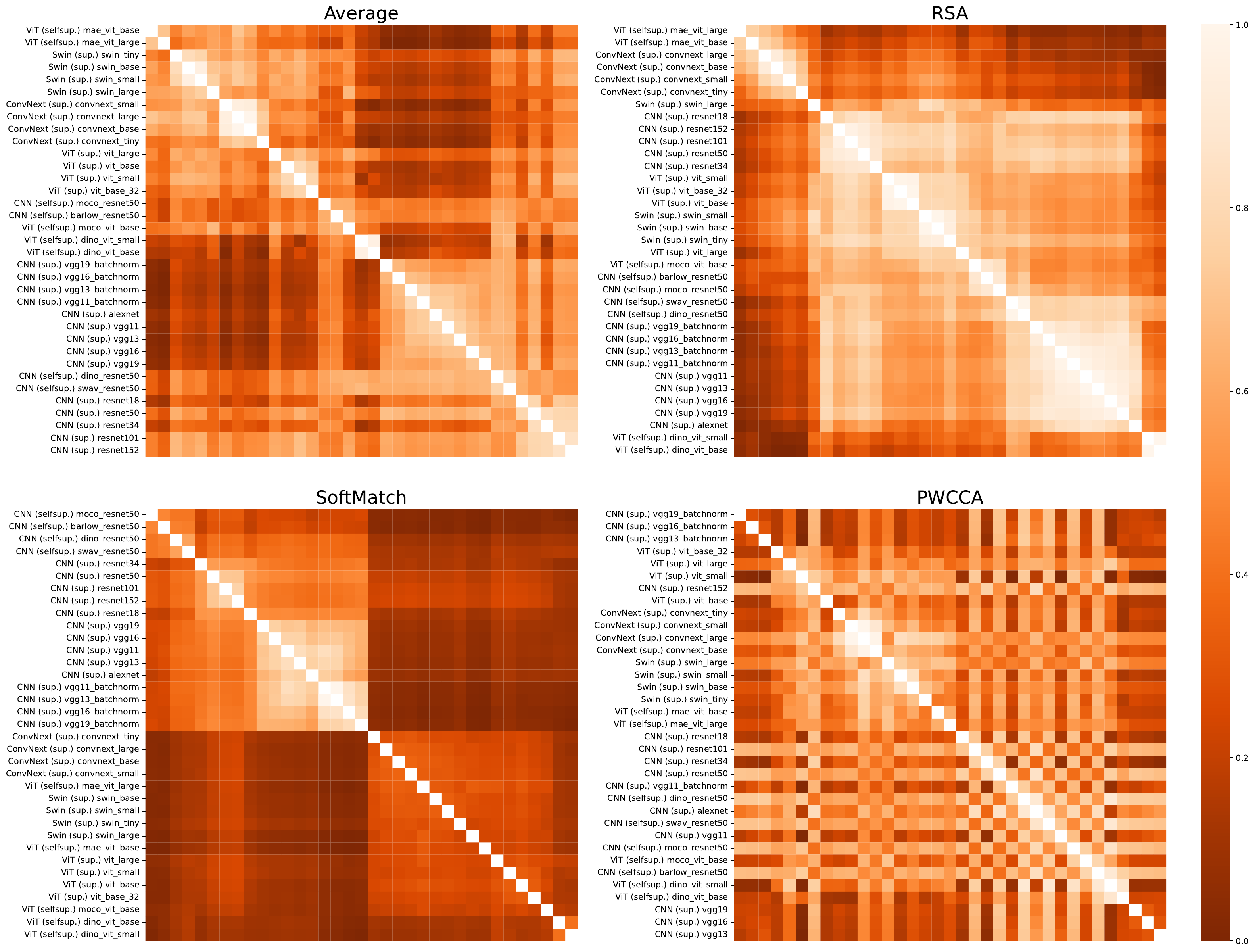}
\caption{Same as \cref{fig:metrics_clustering_imagenet}, but using CIFAR10 instead of ImageNet.}
\label{fig:metrics_clustering_cifar10_1}
\end{figure}

\begin{figure}[htbp!]
\centering
\includegraphics[width=0.95\linewidth]{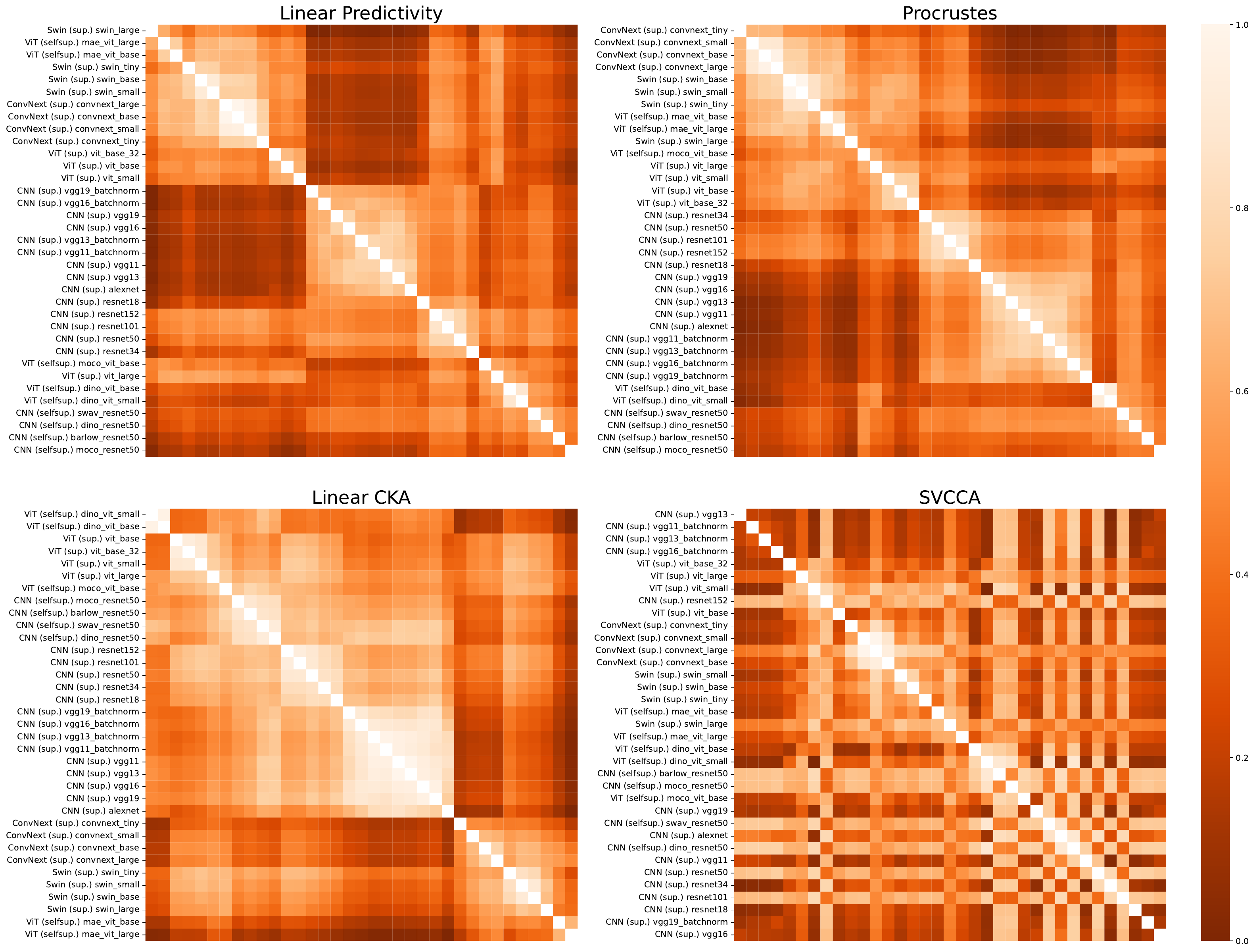}
\caption{Same as \cref{fig:metrics_clustering_imagenet_2}, but using CIFAR10 instead of ImageNet.}
\label{fig:metrics_clustering_cifar10_2}
\end{figure}

\begin{figure}[htbp!]
\centering
\includegraphics[width=\linewidth]{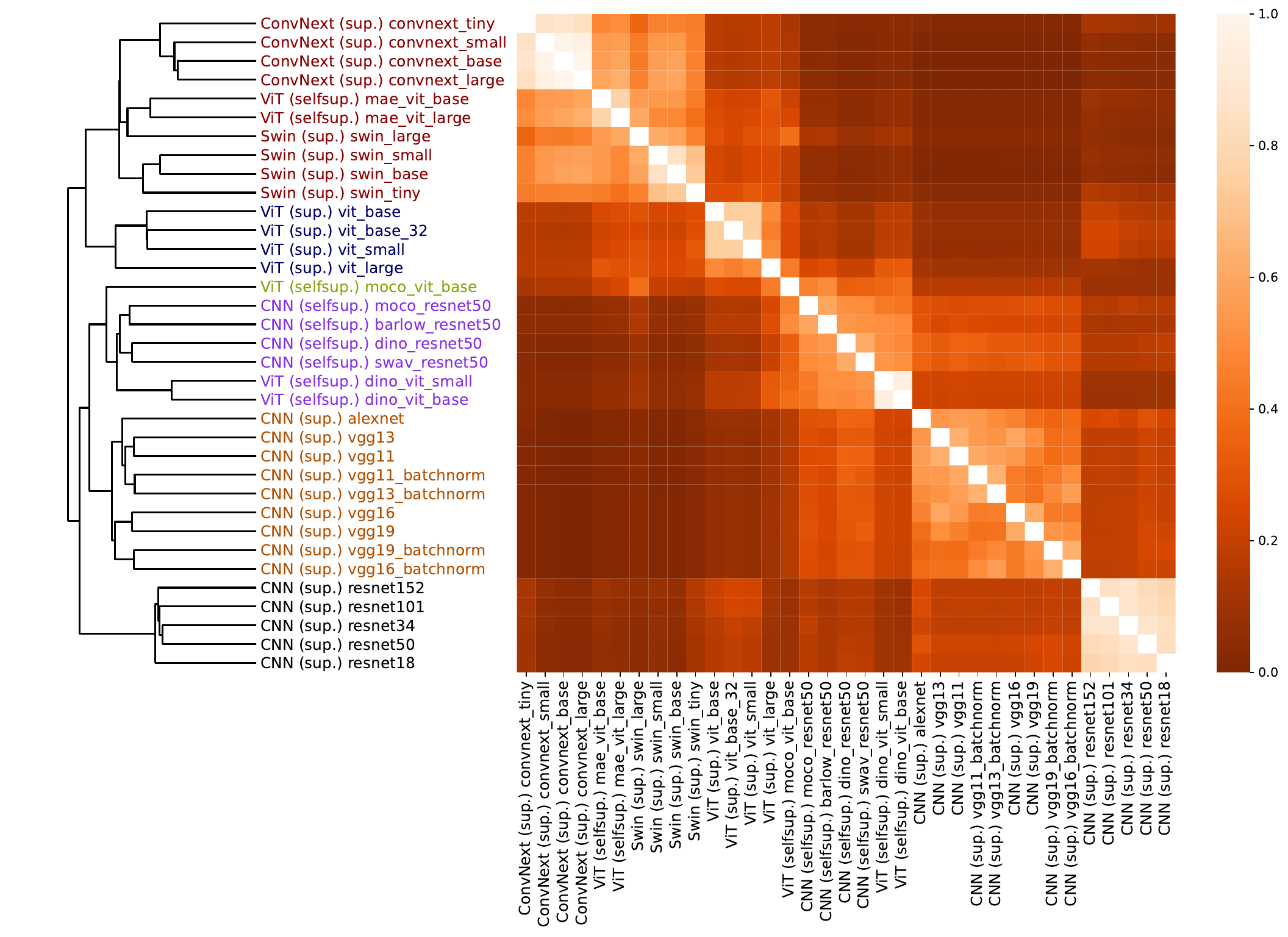}
\caption{Compared to ImageNet in \cref{fig:snf_clustering_ImageNet100}, the clustering result is a little different but similar, demonstrating the SNF results and metric's separability are also influenced by datasets, while also preserving a certain degree of stability. The supervised models are clustered in the same way, but the clusters for the {self-supervised} models changed, demonstrating that the {self-supervised} way leads to a special but kind of unifying representational geometry.}
\label{fig:snf_clustering_cifar10}
\end{figure}

\begin{figure}[htbp!]
\centering
\includegraphics[width=0.95\linewidth]{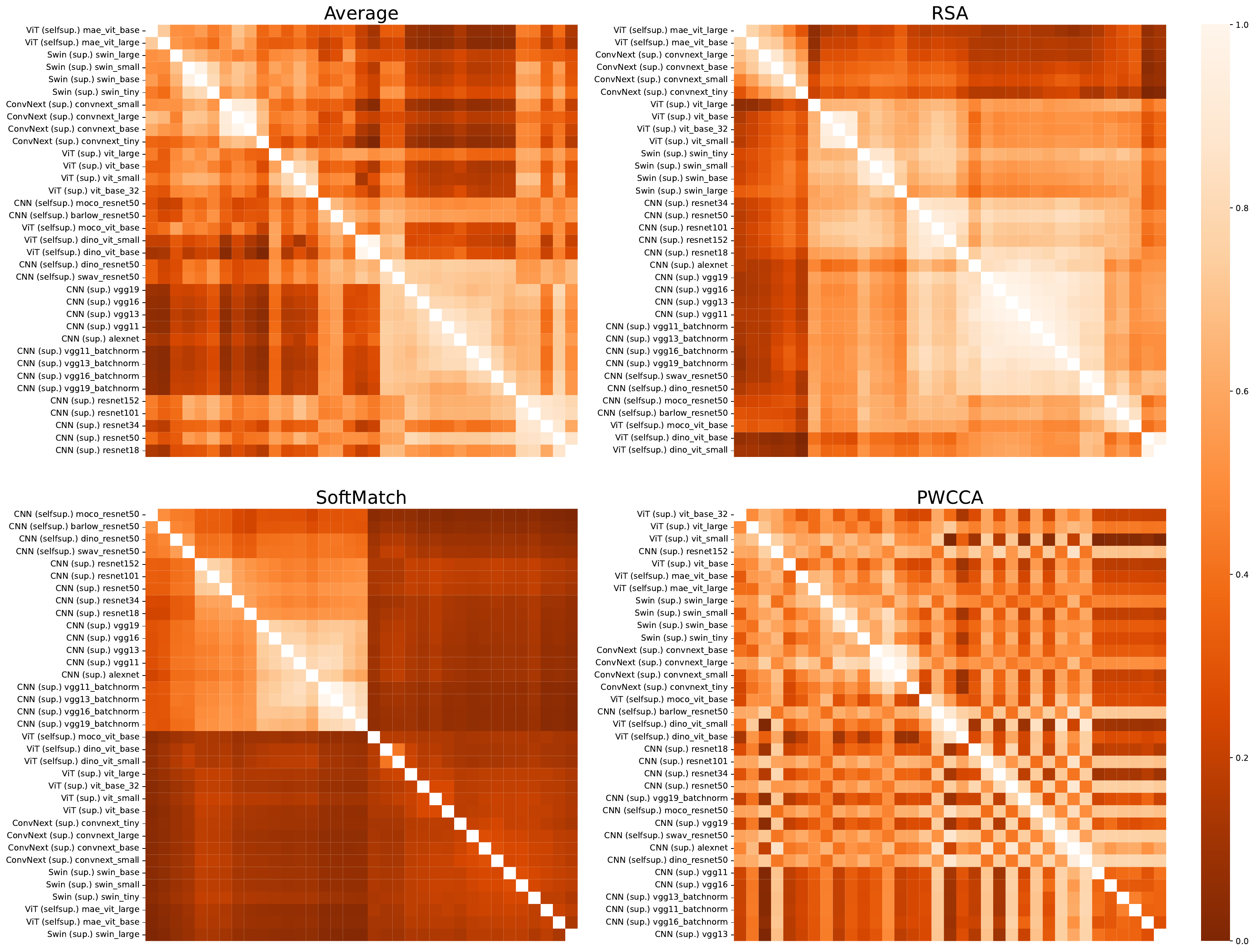}
\caption{Same as \cref{fig:metrics_clustering_imagenet}, but using CIFAR100 instead of ImageNet.}
\label{fig:metrics_clustering_cifar100_1}
\end{figure}

\begin{figure}[htbp!]
\centering
\includegraphics[width=0.95\linewidth]{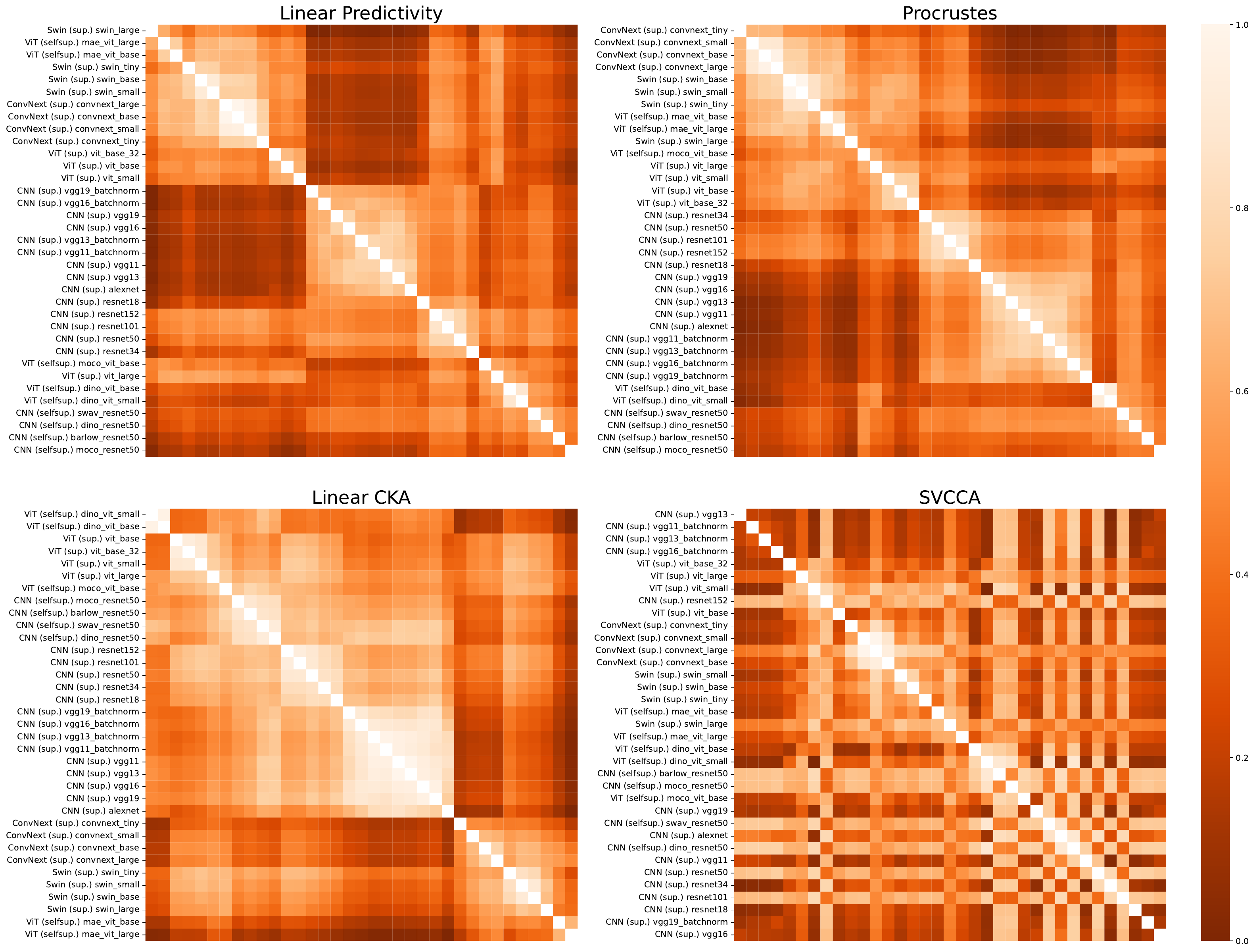}
\caption{Same as \cref{fig:metrics_clustering_imagenet_2}, but using CIFAR100 instead of ImageNet.}
\label{fig:metrics_clustering_cifar100_2}
\end{figure}

\begin{figure}[htbp!]
\centering
\includegraphics[width=\linewidth]{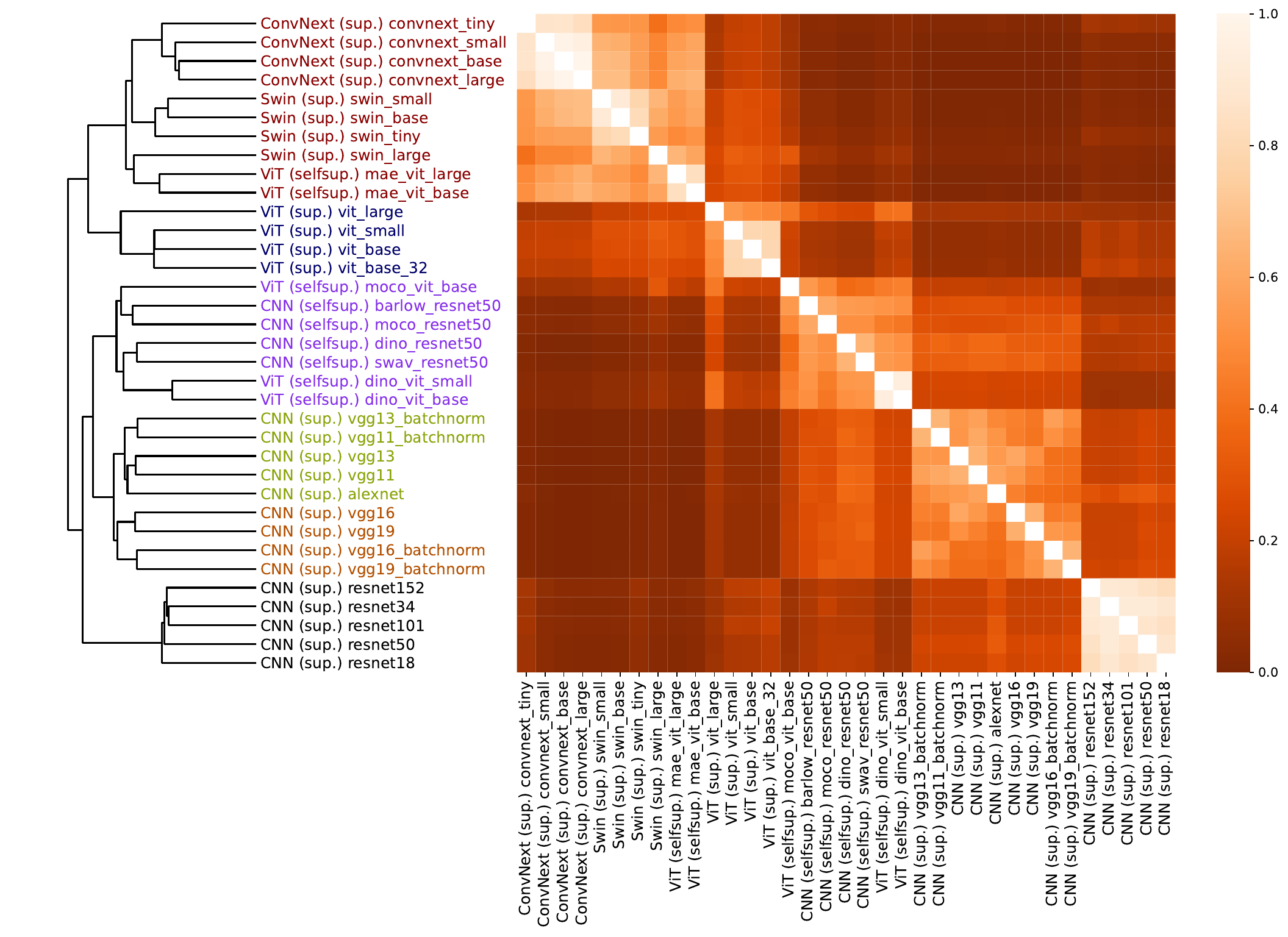}
\caption{Same as \cref{fig:snf_clustering_cifar10}, but using CIFAR100 instead of CIFAR10.}
\label{fig:snf_clustering_cifar100}
\end{figure}

\newpage

\section{Cophenetic Correlation Coefficients for Clustering}

\begin{figure}[htbp!]
\centering
\includegraphics[width=\linewidth]{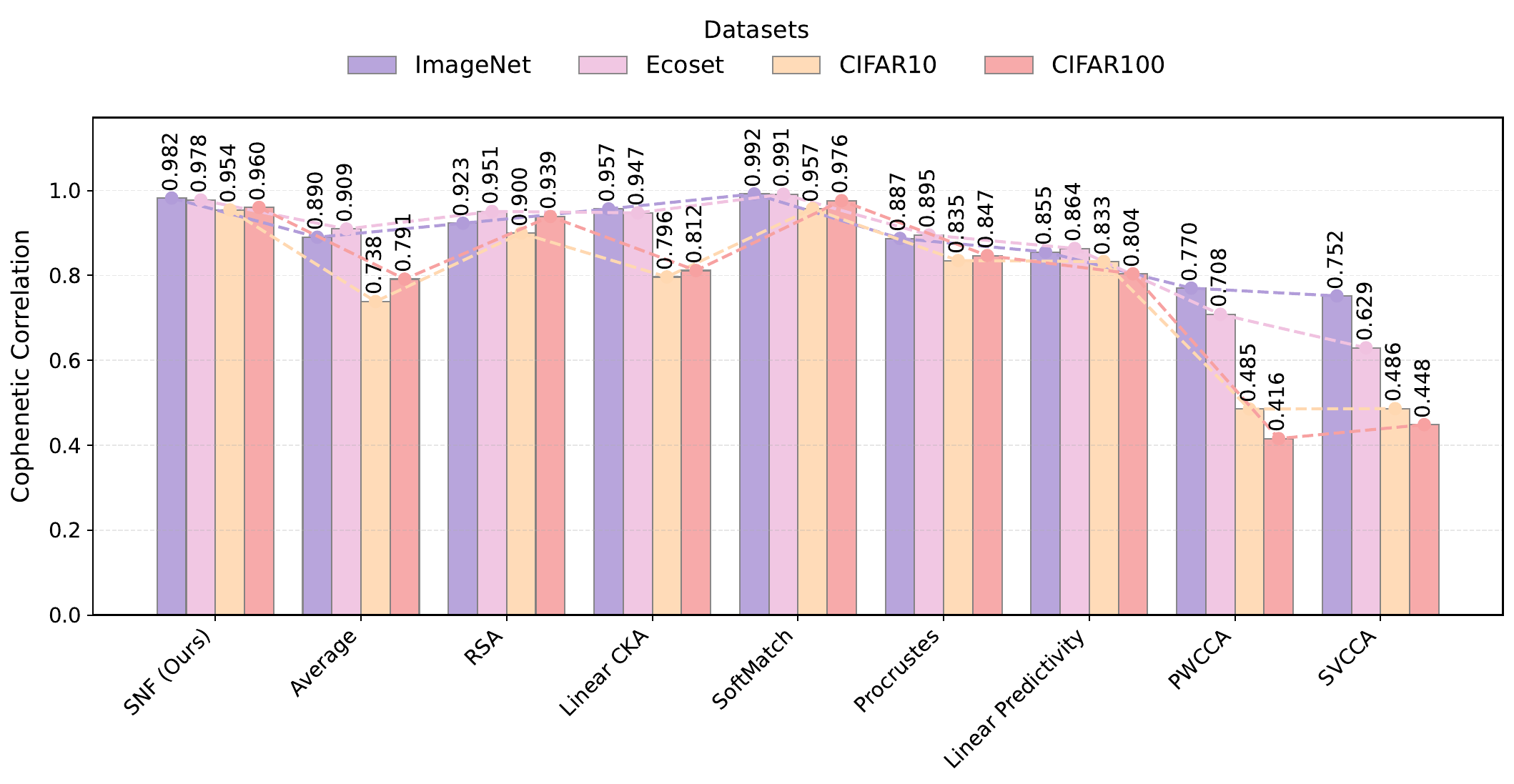}
\caption{Cophenetic correlation coefficients (CCC) for hierarchical clusterings induced by each metric on four datasets. Higher CCC (closer to $1$) means the clustering more faithfully preserves the original pairwise structure. Bars show CCC on ImageNet, Ecoset, CIFAR10, and CIFAR100 (values annotated; lines trace dataset-wise trends).}
\label{fig:cophenetic}
\end{figure}

\newpage
\section{Metrics' Cross Layer Consistency}

For each metric and dataset, we also test whether the similarity structure is stable across depth. For CNNs, if batch normalization layers exist (e.g. ResNet), we count layers by them; if not (e.g., VGG, AlexNet), we count layers by ReLU units. For ViTs, one feature extraction unit (first layer normalization + attention block + second layer normalization) is counted as one layer, and we take the output after the second layer normalization. We select the layer for a normalized depth $d \in (0,1]$ via $\ell=\lfloor dL \rfloor$, where $L$ is the total number of layers. For each metric, one depth corresponds to one matrix. We vectorized the off-diagonal entries, and computed Pearson correlations between the 3 depth pairs. We found that SNF and SoftMatch show the highest depth consistency, whereas the CKA and CCA variants are less stable. Although the representations in different layers could be greatly different, the SNF could still identify the layer-model belonging relationship and the difference between model families.

\begin{figure}[htbp!]
\centering
\includegraphics[width=\linewidth]{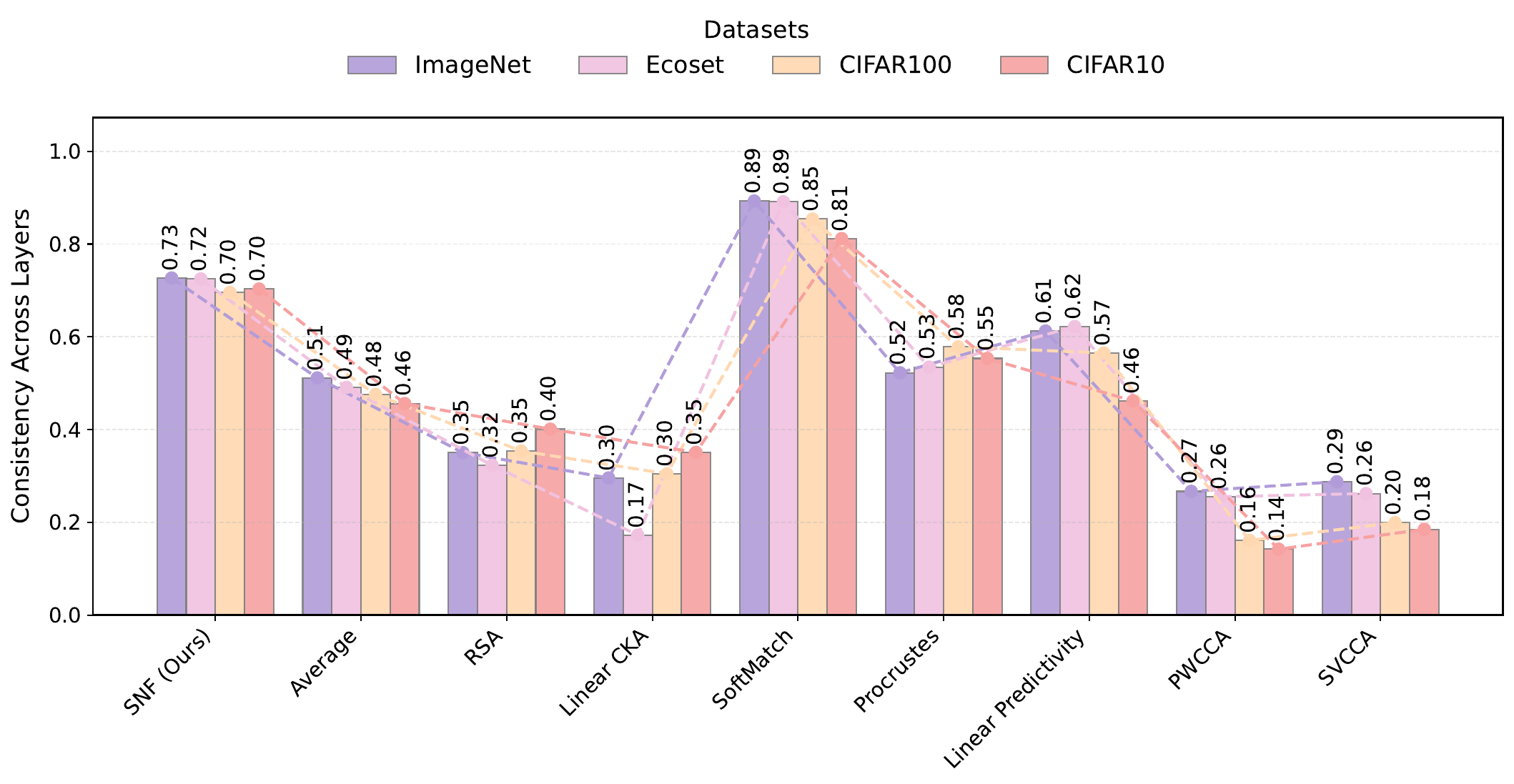}
\caption{Cross-layer consistency of inter-model similarity. The bar height is the mean of the three depth pairs correlations; higher values indicate greater consistency.}
\label{fig:cross_layers}
\end{figure}

\section{The Usage of Large Language Models (LLMs)}
LLMs are mainly used in two ways. For aiding or polishing writing, they are primarily used to identify typos and make the language more aligned with conventions of academic writing. For retrieval and discovery, LLMs with internet access are used to search for related work.

\end{document}